%% file: manuscript.tex
\documentclass[runningheads]{llncs}
\usepackage[T1]{fontenc}
\usepackage{graphicx}
\usepackage{comment}
\usepackage{amsmath,amssymb} %
\usepackage{color}
\usepackage{xcolor}
\usepackage{booktabs}
\usepackage{subcaption}
\usepackage[colorlinks=true]{hyperref}
\usepackage{bm}
\usepackage{xspace}
\usepackage{dsfont}
\usepackage{graphics}

\usepackage{textcomp}
\usepackage{microtype}
\usepackage[all]{nowidow}

\usepackage{algorithm,algorithmic}
\usepackage{wrapfig,epsfig}
\usepackage{mathtools}
\usepackage{dsfont}
\usepackage{graphicx}
\usepackage{nomencl}
\usepackage{multirow}
\usepackage{makecell}
\usepackage{adjustbox}
\usepackage{subcaption}
\usepackage{enumerate}
\usepackage{enumitem}
\usepackage{multirow}
\usepackage{amssymb}
\usepackage[symbol]{footmisc}
\usepackage{colortbl}
\usepackage{diagbox}
\usepackage{fontawesome}

\usepackage{booktabs}
\usepackage{bbding}
\usepackage{pifont}
\usepackage{wasysym}
\usepackage{utfsym}

\usepackage{xcolor}
\definecolor{bestcolor}{HTML}{9898FF}

\definecolor{physicalcolor}{HTML}{E74C3C}    
\definecolor{sensorcolor}{HTML}{3498DB}       
\definecolor{digitalcolor}{HTML}{2ECC71}      
\definecolor{bestresult}{HTML}{D5F5E3}        
\definecolor{secondbest}{HTML}{FEF9E7}        

\usepackage{microtype}
\usepackage[all]{nowidow}

\usepackage{hyperref}
\definecolor{modernCite}{RGB}{0, 160, 140}
\definecolor{modernLink}{RGB}{140, 0, 210}  
\definecolor{modernURL}{RGB}{220, 100, 0}   

\hypersetup{
    colorlinks=true,         
    citecolor=modernCite,   
    linkcolor=modernLink,   
    urlcolor=modernURL,   
    pdftitle={My Document},  
}

\usepackage{tcolorbox}
\tcbuselibrary{minted,skins}
\usepackage{float}

\definecolor{codegreen}{rgb}{0.0, 0.5, 0.4}
\definecolor{codegray}{rgb}{0.5, 0.5, 0.5}
\definecolor{codepurple}{rgb}{0.58, 0, 0.82}
\definecolor{codeblue}{rgb}{0.0, 0.4, 0.7}
\definecolor{codered}{rgb}{0.7, 0.1, 0.1}
\definecolor{backcolour}{rgb}{1.0, 1.0, 1.0}

\usepackage{listings}
\lstdefinestyle{pythonstyle}{
    backgroundcolor=\color{backcolour},
    commentstyle=\color{codegreen},
    keywordstyle=\color{codeblue}\bfseries,
    stringstyle=\color{codered},
    basicstyle=\ttfamily\small,
    breakatwhitespace=false,
    breaklines=true,
    captionpos=t,
    keepspaces=true,
    showspaces=false,
    showstringspaces=false,
    showtabs=false,
    tabsize=4,
    frame=none,
    xleftmargin=2mm,
    xrightmargin=2mm,
    aboveskip=0pt,
    belowskip=0pt,
    emph={self, True, False, None},
    emphstyle=\color{codeblue},
    morekeywords={def, class, return, raise, for, in, if, else, import, from, as, with},
}
\usepackage{cleveref}

\newcommand{\toolbox}{\textbf{FDeID-Toolbox}}
\newcommand{\cmark}{\ding{51}}
\newcommand{\xmark}{\ding{55}}

\usepackage{heatmap}

\usepackage[width=125mm,left=11mm,paperwidth=147mm,height=186mm,top=12mm,paperheight=208mm]{geometry}

\makeatletter
\renewcommand\subsubsection{\@startsection{subsubsection}{4}{\z@}%
  {.5em \@plus1ex \@minus.1ex}%
  {-.5em}%
  {\normalfont\normalsize\bfseries}}
\makeatother

\include{math_commands}

\makeatletter
\DeclareRobustCommand\onedot{\futurelet\@let@token\@onedot}
\def\@onedot{\ifx\@let@token.\else.\null\fi\xspace}

\def\eg{\emph{e.g}\onedot}

\makeatother

\makeatletter
\newcommand{\printfnsymbol}[1]{%
  \textsuperscript{\@fnsymbol{#1}}%
}
\makeatother

\begin{document}
\pagestyle{headings}
\mainmatter

\title{FDeID-Toolbox: Face De-Identification Toolbox}

\titlerunning{FDeID-Toolbox: Face De-Identification Toolbox}
\authorrunning{Wei et al.}
\author{
Hui Wei \and Hao Yu \and Guoying Zhao}
\institute{ELLIS Institute Finland\\Center for Machine Vision and Signal Analysis (CMVS), \\University of Oulu, Finland}

\maketitle

\begin{abstract}
Face de-identification (FDeID) aims to remove personally identifiable information from facial images while preserving task-relevant utility attributes such as age, gender, and expression. 
It is critical for privacy-preserving computer vision, yet the field suffers from fragmented implementations, inconsistent evaluation protocols, and incomparable results across studies. 
These challenges stem from the inherent complexity of the task: FDeID spans multiple downstream applications (\eg, age estimation, gender recognition, expression analysis) and requires evaluation across three dimensions (\eg, privacy protection, utility preservation, and visual quality), making existing codebases difficult to use and extend. 
To address these issues, we present \toolbox, a comprehensive toolbox designed for reproducible FDeID research. 
Our toolbox features a modular architecture comprising four core components: (1)~standardized data loaders for mainstream benchmark datasets, (2)~unified method implementations spanning classical approaches to SOTA generative models, (3)~flexible inference pipelines, and (4)~systematic evaluation protocols covering privacy, utility, and quality metrics. 
Through experiments, we demonstrate that \toolbox\ enables fair and reproducible comparison of diverse FDeID methods under consistent conditions. 
The code is available at \url{https://github.com/infraface/FDeID-Toolbox}.

\end{abstract}

\section{Introduction}
\label{sec:intro}

The proliferation of visual data collection from surveillance systems, social media platforms, and healthcare monitoring has intensified concerns about facial privacy~\cite{hasan2023presentation}. 
Face de-identification (FDeID) addresses this issue by removing personally identifiable information from facial images while preserving task-relevant attributes such as age~\cite{chow2024personalized}, gender~\cite{he2024diff}, and expression~\cite{cai2024disguise}. This capability is essential for enabling privacy-preserving computer vision research and deployment, particularly as regulations like the EU AI Act (Regulation (EU) 2024/1689)~\cite{EUAIAct2024} and HIPAA~\cite{HHSPrivacyRule} impose strict requirements on biometric data processing. 

\begin{figure}[t]
    \centering\small
    \includegraphics[width=1.0\columnwidth]{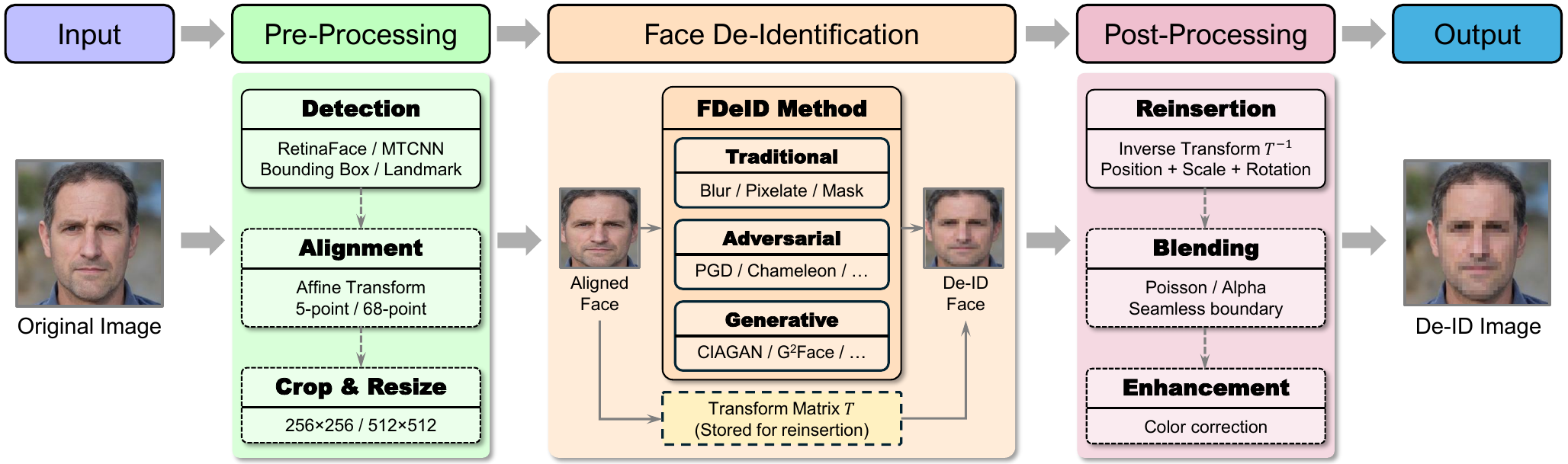}
    \vspace{-5mm}
    \caption{FDeID Pipiline. An example of the components of a FDeID pipeline including pre-processing, face de-identification, and post-processing. Components shown with dashed borders are optional and method-dependent.}
    \label{fig:pipeline}
    \vspace{-5mm}
\end{figure}

Over the past two decades, the FDeID field has witnessed remarkable methodological diversity~\cite{newton2005preserving,gross2006model,du2014garp,hukkelaas2019deepprivacy,cao2021personalized}. Early approaches relied on simple image processing techniques such as blurring and pixelation~\cite{hudson1996techniques,boyle2000effects}, which provide intuitive privacy protection but severely degrade visual utility. Classical methods like $k$-Same~\cite{newton2005preserving} introduced principled privacy guarantees through $k$-anonymity~\cite{sweeney2002k}, yet struggled with naturalness. More recently, adversarial perturbation-based methods~\cite{dong2018boosting,yang2021towards} have emerged as a promising direction, crafting imperceptible noise to fool recognition systems while maintaining visual fidelity. In parallel, generative approaches leveraging GANs~\cite{wu2019privacy} and diffusion models~\cite{park2025facial} have advanced the realism of synthesized faces for FDeID. 
Across these diverse approaches, most FDeID methods share a common processing pipeline consisting of three stages, as shown in \Cref{fig:pipeline}: (i)~optional pre-processing (\eg, face detection, alignment, and cropping); (ii)~the FDeID operation; and (iii)~optional post-processing (\eg, reinsertion, blending, and enhancement). The pre- and post-processing steps are method-dependent and not universally required.

Despite this methodological progress, the field suffers from standardization challenges that hinders scientific progress. Based on our review of literature in the field, we identified three factors that fragment the field and limit the interpretability and reproducibility of reported results. \textbf{First}, utility evaluation vary dramatically across existing FDeID methods~\cite{gross2008semi,maximov2020ciagan,wen2022IdentityMask,wen2023divide}. For example, methods for human–computer interaction emphasize expression recognition~\cite{yang2021systematical,zhai2022a3gan}, healthcare-oriented approaches require preservation of physiological signals such as remote photoplethysmography (rPPG)~\cite{savic2023identification}, and social media applications depend on demographic attributes (e.g., age and gender) for content moderation~\cite{he2024diff,liu20253d}. The absence of a unified framework capable of evaluating multiple utility dimensions simultaneously makes it difficult to assess trade-offs between privacy protection and utility preservation. \textbf{Second}, evaluation protocols differ widely across studies, with inconsistent choices of datasets, recognition models, and metrics, rendering cross-method comparisons unreliable~\cite{kung2025face,zhu2024facemotionpreserve}. \textbf{Third}, existing implementations are scattered across heterogeneous and incompatible codebases, forcing researchers to re-implement methods and evaluation pipelines before conducting comparative studies, thereby impeding cumulative progress. 

Existing toolboxes~(\Cref{tab:comparison}) address these issues only partially.
DeepPrivacy~\cite{hukkelaas2019deepprivacy} and DeepPrivacy2~\cite{hukkelaas2023deepprivacy2} provide training for their own generative methods but support no other paradigms and evaluate no downstream utility.
The lightweight \texttt{deface}~\cite{drawitsch2020deface} tool targets video anonymization via traditional methods alone, providing neither training nor evaluation.
None of these tools implements a unified interface across method paradigms, supports more than one dataset, or evaluates more than a single downstream task.

To address these challenges, we introduce \toolbox, a comprehensive toolbox designed as shared infrastructure for reproducible FDeID research. \toolbox~adopts a modular architecture with four core components: standardized data loaders for widely used benchmark datasets; unified implementations of representative FDeID methods spanning classical techniques, adversarial perturbations, and state-of-the-art generative models; flexible inference pipelines; and systematic evaluation protocols covering privacy protection, utility preservation, and visual quality. 
In total, the toolbox includes 17 representative FDeID methods exposed through a consistent \texttt{BaseDeidentifier} (see \Cref{alg:base_interface}) interface and supports extensible data loaders with standardized preprocessing. Evaluation is performed across privacy metrics, utility measures for six attributes (age, gender, ethnicity, landmarks, expression, and rPPG), and visual quality indicators, enabling fair and comprehensive comparison. All experiments are configured through YAML files, ensuring transparency and reproducibility. Together, these features position \toolbox~as a unified platform for rigorous, reproducible, and extensible research in FDeID.

\section{Related Work}

\begin{table}[t]
    \centering
    \caption{Comparison of FDeID Toolboxes. Existing tools focus on specific method families and lack comprehensive utility evaluation, while our FDeID-Toolbox provides unified infrastructure spanning various method paradigms.}
    \vspace{1mm}
    \label{tab:comparison}
    \setlength{\tabcolsep}{5pt}
    \resizebox{\textwidth}{!}{
    \begin{tabular}{lcccccccc}
    \toprule
    \multirow{2}{*}{\textbf{Toolbox}} & \multirow{2}{*}{\textbf{\makecell[c]{Dataset\\ Support}}} & \multirow{2}{*}{\textbf{\makecell[c]{Training\\ Support}}} & \multicolumn{5}{c}{\makecell[c]{\textbf{Utility Preservation Evaluation}}} \\
    \cmidrule(r){4-9}
    & & & Age & Gender & Ethnicity & Landmark & Expression & rPPG \\
    \midrule
       DeepPrivacy~\cite{hukkelaas2019deepprivacy} & \cmark & \cmark & \xmark & \xmark & \xmark & \xmark & \xmark & \xmark \\
       DeepPrivacy2~\cite{hukkelaas2023deepprivacy2} & \cmark & \cmark & \xmark & \xmark & \xmark & \xmark & \xmark & \xmark \\
       deface~\cite{drawitsch2020deface} & \xmark & \xmark & \xmark & \xmark & \xmark & \xmark & \xmark & \xmark \\
       FDeID-Toolbox (Ours)  & \cmark & \cmark & \cmark & \cmark & \cmark & \cmark & \cmark & \cmark \\
    \bottomrule
    \end{tabular}
    }
    \vspace{-5mm}
\end{table}

\vspace{-3mm}
\subsection{Face De-Identification}
FDeID aims to remove personally identifiable information from facial images while preserving task-relevant attributes. Early approaches employed handcrafted image processing operations such as Gaussian blur, pixelation, and masking~\cite{hudson1996techniques,boyle2000effects}, which provide intuitive privacy protection but suffer from fundamental trade-offs between privacy and utility preservation. Formal privacy guarantees were introduced through $k$-anonymity~\cite{sweeney2002k} based methods: the $k$-Same algorithm~\cite{newton2005preserving} clusters faces and replaces each with the cluster average, theoretically bounding recognition accuracy at $1/k$, with subsequent refinements addressing utility preservation~\cite{gross2005integrating,gross2006model,meng2014face}. Adversarial perturbation-based methods craft imperceptible noise to mislead face recognition systems~\cite{shamshad2023clip2protect,NEURIPS2022_dccbeb7a,yang2021towards}, achieving near-perfect visual fidelity but facing transferability challenges across black-box settings. More recently, generative model-based approaches have emerged as the dominant paradigm: GAN-based methods~\cite{wu2019privacy,maximov2020ciagan,zhang2024rbgan} synthesize photorealistic de-identified faces through encoder-decoder architectures with identity-adversarial losses, while diffusion-based methods~\cite{kung2025face,park2025facial} achieve superior visual quality through iterative denoising. Specialized variants address reversibility for authorized recovery~\cite{wen2022IdentityMask,yang2024g}, attribute preservation~\cite{zhai2022a3gan,meden2023face}, and domain-specific utility requirements such as rPPG signal preservation~\cite{savic2023identification}. Despite this methodological richness, the field suffers from fragmented implementations across incompatible codebases, inconsistent evaluation metrics, and narrow utility assessments. These challenges motivate the development of the \toolbox.

\vspace{-5mm}
\begin{algorithm}[h]
\caption{Base De-identifier Interface}
\label{alg:base_interface}
\begin{adjustbox}{max width=0.75\textwidth}
\begin{lstlisting}[language=Python]
# All methods inherit from BaseDeidentifier
class BaseDeidentifier:
    def __init__(self, config):
        self.config = config
        self.device = config.get("device", "cuda")
    
    def deidentify(self, face_img):
        """Apply de-identification to aligned face."""
        raise NotImplementedError

# Example usage
config = {"method": "CIAGAN", "device": "cuda"}
deid = BaseDeidentifier(config)
result = deid.deidentify(aligned_face)  # [H, W, C]
\end{lstlisting}
\end{adjustbox}
\end{algorithm}
\vspace{-10mm}

\subsection{Existing Tools}
As shown in \Cref{tab:comparison}, several open-source tools have been developed for the FDeID task, yet each addresses only a narrow subset of research requirements. DeepPrivacy~\cite{hukkelaas2019deepprivacy} pioneered GAN-based FDeID with a conditional inpainting approach, later extended in DeepPrivacy2~\cite{hukkelaas2023deepprivacy2} to support full-body de-identification at higher resolution. While these tools provide training infrastructure for their specific generative methods, they lack multiple dataset loaders and do not evaluate utility preservation across downstream tasks. 
The deface~\cite{drawitsch2020deface} tool offers a lightweight command-line interface for video anonymization using traditional methods (blur, pixelation, masking), but provides no training capability, evaluation framework, or support for learning-based methods. 
Critically, none of these tools support comprehensive evaluation across privacy, utility, and quality dimensions. Our \toolbox~addresses these limitations by unifying representative methods under a common interface, providing data loaders for mainstream datasets, and implementing systematic evaluations covering six utility attributes alongside privacy and quality metrics.

\section{The \toolbox}
\label{sec:toolbox}

\subsection{Architecture Overview}
\toolbox\ is built around three design principles that directly address the standardization gaps identified in \Cref{sec:intro}.
\textit{Modularity}: each component exposes a well-defined interface so that datasets, methods, and metrics can be swapped or extended without touching other parts of the codebase.
\textit{Reproducibility}: every experiment is fully specified by a single YAML file covering dataset paths, method hyperparameters, model
checkpoints, and evaluation criteria, making results shareable and auditable.
\textit{Comprehensiveness}: rather than optimizing for a single method family or evaluation dimension, the toolbox covers all three paradigms (traditional, adversarial, generative) and all three evaluation dimensions (privacy, utility, quality).

\begin{figure}[t]
    \centering\small
    \includegraphics[width=1.0\columnwidth]{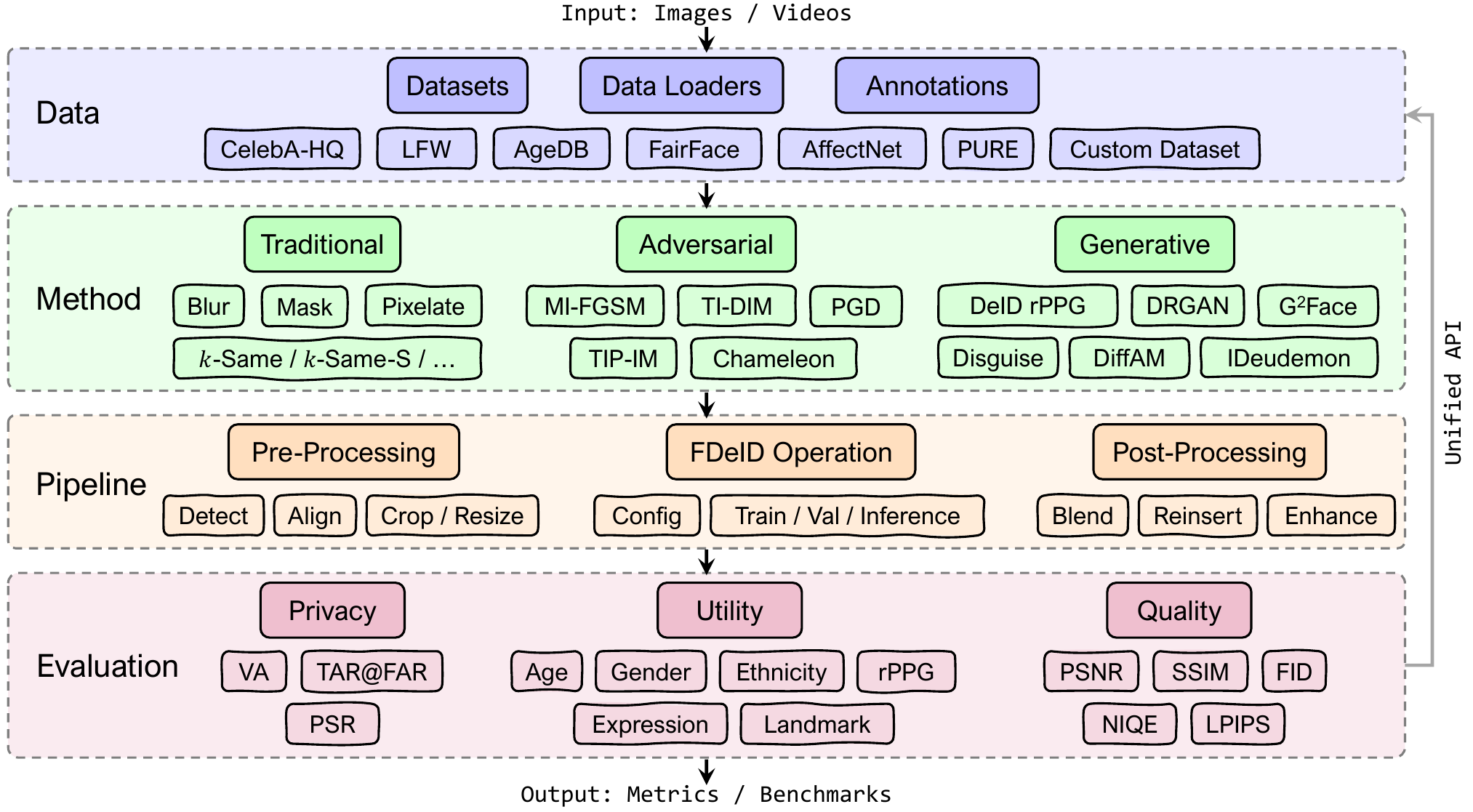}
    \vspace{-6mm}
    \caption{Architecture of the \toolbox. The modular design comprises four core modules: \textcolor{blue!70}{\textbf{Data}} for unified dataset management, \textcolor{green!50!black}{\textbf{Methods}} spanning traditional, adversarial, and generative approaches, \textcolor{orange!70!black}{\textbf{Pipeline}} for pre-processing, training, and post-processing, and \textcolor{purple!70}{\textbf{Evaluation}} integrating privacy, utility, and quality metrics. Components communicate through a unified API, enabling plug-and-play extensibility.}
    \label{fig:arch}
    \vspace{-6mm}
\end{figure}

As shown in \Cref{fig:arch}, \toolbox\ comprises four tightly integrated modules. The \textbf{Data Module} provides standardised data loaders and preprocessing pipelines for six benchmark datasets, exposing a common \texttt{BaseDataset} interface that supports custom datasets with minimal boilerplate.
The \textbf{Method Module} implements 17 FDeID algorithms under a shared \texttt{BaseDeidentifier} interface~(\Cref{alg:base_interface}),
ensuring that every method can be invoked, benchmarked, and composed identically regardless of its underlying paradigm.
The \textbf{Pipeline Module} orchestrates the complete FDeID workflow with each stage independently configurable and optional.
The \textbf{Evaluation Module} unifies privacy, utility, and quality assessment into a single pass, computing all metrics from the same
de-identified outputs under identical conditions.

\vspace{-2mm}
\subsection{Supported Datasets}
Dataset selection is driven by coverage across the three evaluation dimensions and the six utility attributes.
The data module currently supports the following datasets: 1) \textbf{CelebA-HQ}~\cite{karras2018progressive} contains 30,000 high-quality face images at $1024\times1024$ resolution, selected from CelebA~\cite{liu2015deep} and annotated with 40 binary attributes (\eg, \textit{Smiling}, \textit{Eyeglasses}, \textit{Young}). 
We specifically leverage it for {landmark} utility evaluation.
2) \textbf{LFW}~\cite{huang2008labeled} comprises 13,233 unconstrained face images of 5,749 identities collected from the web and is used for privacy and visual quality evaluation. 3) \textbf{AgeDB}~\cite{moschoglou2017agedb} includes 16,516 in-the-wild face images of 570 identities with age annotations and is used for both privacy and age-related utility evaluation. 4) \textbf{FairFace}~\cite{karkkainen2021fairface} consists of 108,501 images with balanced annotations for ethnicity, gender, and age, and is employed for utility evaluation of gender and ethnicity. 5) \textbf{AffectNet}~\cite{mollahosseini2017affectnet} contains over one million facial images annotated with expression labels and is used for expression-related utility evaluation. 6) \textbf{PURE}~\cite{stricker2014non} comprises recordings of 10 identities captured under six different setups, resulting in 60 one-minute video sequences, and is used for evaluating rPPG preservation.

\subsection{Implemented Methods}

\toolbox\ implements 17 representative FDeID methods spanning three technical paradigms. \textit{Traditional methods} achieve privacy through explicit visual degradation or face substitution without learned models. \textit{Adversarial methods} craft imperceptible perturbations that fool face recognition systems while leaving the image visually intact. \textit{Generative methods} synthesize a new facial appearance via learned generative models, offering the greatest flexibility in balancing privacy with visual realism. 

\textbf{Traditional Methods} implement naive image processing techniques and the k-Same family, requiring no training. Naive methods include \emph{Blur}, \emph{Pixelation}, and \emph{Mask}, which serve as important baselines and remain widely deployed in practice despite their limitations. The \emph{k-Same} family replaces each face with a combination of its $k$ nearest neighbors in embedding space: \emph{k-Same}~\cite{newton2005preserving} substitutes a face with the pixel-wise mean of $k$ neighbors; \emph{k-Same-Select}~\cite{gross2005integrating} selects a single representative from the $k$ candidates; and \emph{k-Same-Furthest}~\cite{meng2014face} maximizes privacy by choosing the most dissimilar face within the $k$-neighbor set.

\textbf{Adversarial Methods} craft imperceptible perturbations to fool face recognition systems. Our toolbox includes \emph{MI-FGSM}~\cite{dong2018boosting}, \emph{PGD}~\cite{madry2018towards}, \emph{TI-DIM}~\cite{dong2019evading}, \emph{TIP-IM}~\cite{yang2021towards}, and \emph{Chameleon}~\cite{chow2024personalized}. These methods optimize adversarial perturbation $\delta$ to maximize recognition loss while constraining $\|\delta\|_\infty \leq \epsilon$. 

\textbf{Generative Methods} synthesize de-identified face images using learned generative models. \emph{CIAGAN}~\cite{maximov2020ciagan} employs a conditional GAN with an identity-guided discriminator to generate de-identified faces. \emph{Adv-Makeup}~\cite{yin2021adv} transfers adversarial makeup patterns onto facial regions to mislead recognition systems while maintaining visually plausible appearances. \emph{AMT-GAN}~\cite{hu2022protecting} extends adversarial makeup transfer by incorporating feature disentanglement to improve visual quality. \emph{DeID-rPPG}~\cite{savic2023identification} performs video-based FDeID while preserving rPPG signals for health monitoring.  \emph{G$^{2}$Face}~\cite{yang2024g} enables high-fidelity, reversible face de-identification, allowing authorized re-identification when required.
\emph{WeakenDiff}~\cite{salar2025enhancing} leverages diffusion models to adversarially modify latent representations for identity suppression.

\subsection{Pre-Processing, FDeID Operation, and Post-Processing}

\toolbox\ provides end-to-end pipelines standardizing the complete FDe-ID workflow. As shown in \Cref{fig:pipeline}, the pipeline comprises three independently configurable stages; pre- and post-processing are optional and method-dependent.

\noindent\textbf{Pre-Processing.}
For methods operating on cropped face regions, raw images are transformed into standardized inputs via three steps: (i)~\textit{face detection} using RetinaFace~\cite{deng2020retinaface}, yielding bounding boxes and facial landmarks; (ii)~\textit{alignment} via an affine transformation computed from the detected landmarks, with the transform matrix $T$ stored for later reinsertion; and (iii)~\textit{cropping and resizing} to a standardized resolution.

\noindent\textbf{FDeID Operation.}
All methods are configured via YAML files specifying hyperparameters, model paths, and evaluation settings. Traditional and adversarial methods require no training; generative methods implement a \texttt{BaseTrainer} interface supporting GAN-based and diffusion-based paradigms with checkpointing and loss logging. At inference, all methods expose a unified \texttt{BaseDeidentifier} interface dispatched via \texttt{get\_deidentifier(config)}.

\noindent\textbf{Post-Processing.} For methods operating on cropped regions, the de-identified face is reintegrated into the original image via: (i)~\textit{reinsertion} using the inverse transform $T^{-1}$ encoding position, scale, and rotation; (ii)~\textit{blending} to eliminate boundary artifacts; and (iii)~optional \textit{enhancement} to reconcile chromatic inconsistencies with the surrounding image.

\section{Benchmarking and Analysis}
\label{sec:exp}

\subsection{Metrics}

Following the mainstream methods~\cite{maximov2020ciagan,chow2024personalized}, we evaluate FDeID methods across three dimensions: privacy, utility, and quality. 

\noindent\textbf{Privacy Metrics.} We measure FDeID effectiveness using face verification accuracy on multiple recognition models and report: (1) \emph{Verification Accuracy} (VA$\downarrow$): the rate at which de-identified faces are correctly matched to originals at a fixed threshold; (2) \emph{True Accept Rate at 0.1\% FAR} (TAR@FAR=0.1\%$\downarrow$): verification rate under strict security settings; and (3) \emph{Protection Success Rate} (PSR$\uparrow$): also known as Attack Success Rate (ASR), percentage of faces successfully de-identified below the recognition threshold.

\noindent\textbf{Utility Metrics.} We assess utility preservation using pre-trained classifiers for age (Mean Absolute Error, MAE$\downarrow$), gender (Accuracy, Acc$\uparrow$), ethnicity (Accuracy, Acc$\uparrow$), expression (Accuracy, Acc$\uparrow$), and landmark detection (Normalized Mean Error, NME$\downarrow$). For rPPG preservation, we measure Mean Absolute Error of estimated heart rate (HR-MAE$\downarrow$).

\noindent\textbf{Quality Metrics.} 
Visual quality, in the context of FDeID, refers to the degree to which a de-identified image preserves the original image's appearance (\eg, pixel structure, perceptual texture, distribution) independent of identity. This source-fidelity framing is distinct from assessing whether a synthesized face looks photorealistic as a new identity. We evaluate source fidelity through five complementary metrics. \textit{PSNR}$\uparrow$ and \textit{SSIM}$\uparrow$~\cite{wang2004image} measure pixel-level fidelity. \textit{LPIPS}$\downarrow$~\cite{zhang2018unreasonable} quantifies perceptual distance using deep features. \textit{FID}$\downarrow$~\cite{heusel2017gans} assesses distributional similarity between de-identified and real face sets via Inception-v3~\cite{szegedy2016rethinking} features, capturing overall realism. \textit{NIQE}$\downarrow$~\cite{mittal2012making} provides reference-free quality assessment by measuring deviation from natural scene statistics.

\begin{figure}[t]
    \centering\small
    \includegraphics[width=1.0\columnwidth]{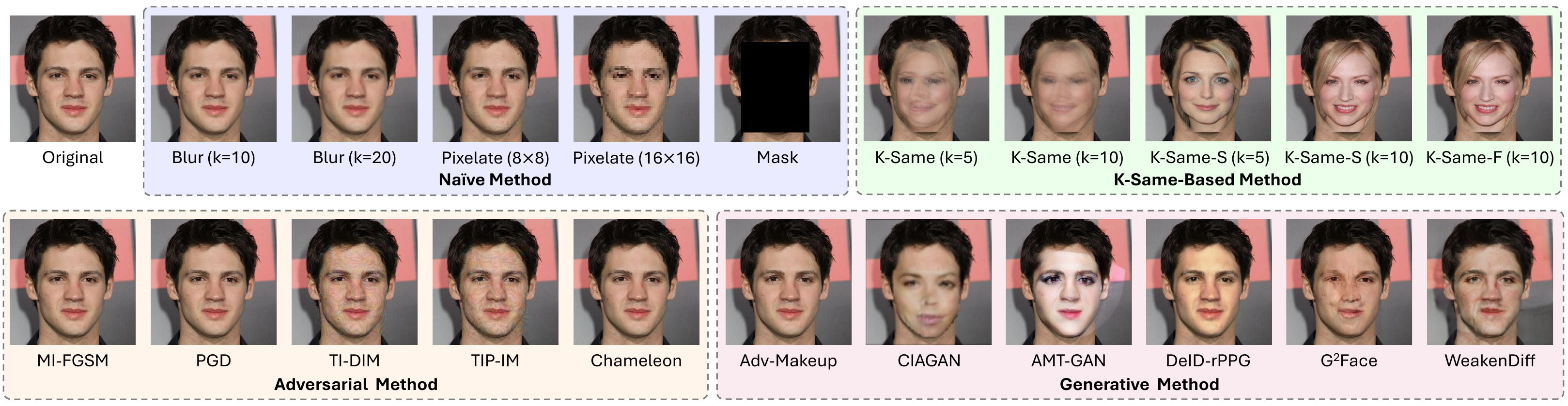}
    \vspace{-6mm}
    \caption{\textbf{Qualitative comparison of FDeID methods on a sample face image}. Traditional naive methods (blur, pixelate, mask, k-Same variants~\cite{newton2005preserving,gross2005integrating,meng2014face}) provide privacy but degrade visual quality. Adversarial methods (MI-FGSM~\cite{dong2018boosting}, PGD~\cite{madry2018towards}, TI-DIM~\cite{dong2019evading}, TIP-IM~\cite{yang2021towards}, Chameleon~\cite{chow2024personalized}) focus on introducing visually imperceptible perturbations. Generative methods (Adv-Makeup~\cite{yin2021adv}, CIAGAN~\cite{maximov2020ciagan}, AMT-GAN~\cite{hu2022protecting}, DeID-rPPG~\cite{savic2023identification}, G$^{2}$Face~\cite{yang2024g}, WeakenDiff~\cite{salar2025enhancing}) synthesize de-identified faces with varying degrees of identity change and visual realism.}
    \label{fig:show}
    \vspace{-6mm}
\end{figure}

\subsection{Implementation Details}
We use RetinaFace~\cite{deng2020retinaface} with five-point landmarks for face detection and alignment. Age prediction is evaluated on AgeDB~\cite{moschoglou2017agedb}, while gender and ethnicity classification are conducted on FairFace~\cite{karkkainen2021fairface} using $224\times224$ inputs, covering two gender classes and seven ethnicity categories. Expression recognition employs POSTER~\cite{zheng2023poster} with $224\times224$ inputs and seven expression classes. Facial landmark detection is performed using HRNet~\cite{wang2020deep}, which predicts 29 keypoints from $256\times256$ inputs. For rPPG preservation, we adopt FactorizePhys~\cite{joshi2024factorizephys}, a video-based model with 3D convolutional blocks and factorized self-attention based on non-negative matrix factorization, operating on video clips of 160 frames at a resolution of $72\times72$. For adversarial-based methods, we set the perturbation budget to $\epsilon{=}8/255$, the step size to $\alpha{=}2/255$, and the number of iterations to 20 under an $\ell_\infty$ norm constraint. All experiments are conducted on the supercomputer using AMD MI250X GPUs with ROCm~6.2.2 and PyTorch~2.5.1.

\begin{table}[t]
    \centering
    \caption{\textbf{Privacy protection performance on LFW~\cite{huang2008labeled} and AgeDB~\cite{moschoglou2017agedb}.} We report verification accuracy (VA$\downarrow$), TAR@FAR=0.1\%$\downarrow$, and Protection Success Rate (PSR$\uparrow$) across three recognition models (ArcFace~\cite{deng2019arcface}, CosFace~\cite{wang2018cosface}, AdaFace~\cite{kim2022adaface}). Darker background colors indicate better performance. All results are from our unified re-implementation under identical conditions.}
    \label{tab:privacy}
    \setlength{\tabcolsep}{2pt}
    \resizebox{\textwidth}{!}{
    \begin{tabular}{l|c|ccc|ccc|ccc|ccc|ccc|ccc}
    \toprule
    & & \multicolumn{9}{c|}{\textbf{LFW}} & \multicolumn{9}{c}{\textbf{AgeDB}} \\
    \cmidrule(lr){3-11} \cmidrule(lr){12-20}
    \multirow{2}{*}{\textbf{Method}} & \multirow{2}{*}{\textbf{Type}} & \multicolumn{3}{c|}{\textbf{ArcFace-R100}} & \multicolumn{3}{c|}{\textbf{CosFace-R50}} & \multicolumn{3}{c|}{\textbf{AdaFace-R50}} & \multicolumn{3}{c|}{\textbf{ArcFace-R100}} & \multicolumn{3}{c|}{\textbf{CosFace-R50}} & \multicolumn{3}{c}{\textbf{AdaFace-R50}} \\
    \cmidrule(lr){3-5} \cmidrule(lr){6-8} \cmidrule(lr){9-11} \cmidrule(lr){12-14} \cmidrule(lr){15-17} \cmidrule(lr){18-20}
    & & VA$\downarrow$ & TAR$\downarrow$ & PSR$\uparrow$ & VA$\downarrow$ & TAR$\downarrow$ & PSR$\uparrow$ & VA$\downarrow$ & TAR$\downarrow$ & PSR$\uparrow$ & VA$\downarrow$ & TAR$\downarrow$ & PSR$\uparrow$ & VA$\downarrow$ & TAR$\downarrow$ & PSR$\uparrow$ & VA$\downarrow$ & TAR$\downarrow$ & PSR$\uparrow$ \\
    \midrule
    \rowcolor{gray!10}
    Original (No Face De-ID) & -- & 99.25 & 98.37 & -- & 99.23 & 98.33 & -- & 99.23 & 98.40 & -- & 98.17 & 95.82 & -- & 98.05 & 95.62 & -- & 97.71 & 93.57 & -- \\
    \midrule
    Gaussian Blur ($\sigma$=10) & Naive & \va{99.22} & \tar{98.33} & \, \psr{0.40} & \va{99.26} & \tar{98.47} & \, \psr{0.40} & \va{98.97} & \tar{97.83} & \, \psr{0.73} & \va{96.31} & \tar{86.57} & \, \psr{5.58} & \va{96.29} & \tar{84.03} & \, \psr{5.41} & \va{95.77} & \tar{83.03} & \, \psr{4.18} \\  
    Gaussian Blur ($\sigma$=20) & Naive & \va{98.23} & \tar{93.50} & \, \psr{3.37} & \va{98.25} & \tar{94.60} & \, \psr{2.53} & \va{97.37} & \tar{84.80} & \, \psr{3.27} & \va{91.14} & \tar{48.63} & \psr{22.57} & \va{91.50} & \tar{57.65} & \psr{20.19} & \va{89.84} & \tar{30.81} & \psr{16.97} \\ 
    Pixelation (8$\times$8) & Naive & \va{78.11} & \, \tar{4.07} & \psr{29.35} & \va{80.45} & \, \tar{3.24} & \psr{38.36} & \va{74.06} & \, \tar{0.45} & \psr{42.25} & \va{82.43} & \, \tar{4.11} & \psr{45.79} & \va{84.38} & \, \tar{2.54} & \psr{39.73} & \va{84.28} & \tar{28.38} & \psr{32.88} \\ 
    Pixelation (16$\times$16) & Naive & \va{60.29} & \, \tar{1.45} & \psr{21.74} & \va{63.97} & \, \tar{4.35} & \psr{10.14} & \va{61.03} & \, \tar{1.45} & \, \psr{8.70} & \va{93.33} & \, \tar{0.00} & \psr{27.27} & \va{80.00} & \, \tar{0.00} & \psr{27.27} & \va{86.67} & \, \tar{0.00} & \, \psr{9.09} \\ 
    Black Mask & Naive & \va{83.33} & \tar{42.86} & \psr{57.14} & \va{83.33} & \tar{57.14} & \psr{35.71} & \va{83.33} & \tar{28.57} & \psr{57.14} & \va{59.29} & \, \tar{0.30} & \, \psr{6.60} & \va{56.15} & \, \tar{0.43} & \, \psr{0.03} & \va{57.27} & \, \tar{0.63} & \, \psr{9.94} \\ 
    \midrule
    $k$-Same ($k$=5) & Class. & \va{65.58} & \, \tar{3.50} & \psr{70.73} & \va{67.72} & \, \tar{3.37} & \psr{56.80} & \va{65.97} & \, \tar{2.87} & \psr{51.93} & \va{56.87} & \, \tar{0.93} & \psr{76.65} & \va{57.60} & \, \tar{0.87} & \psr{65.91} & \va{57.20} & \, \tar{0.50} & \psr{49.87} \\
    $k$-Same ($k$=10) & Class. & \va{68.52} & \, \tar{3.47} & \psr{54.53} & \va{70.34} & \, \tar{4.40} & \psr{37.37} & \va{67.74} & \, \tar{4.17} & \psr{33.63} & \va{57.35} & \, \tar{0.73} & \psr{64.81} & \va{58.05} & \, \tar{1.13} & \psr{47.63} & \va{57.34} & \, \tar{1.33} & \psr{30.89}\\
    $k$-Same-Select ($k$=5) & Class. & \va{57.58} & \, \tar{0.73} & \psr{94.73} & \va{57.86} & \, \tar{0.93} & \psr{93.73} & \va{58.12} & \, \tar{0.63} & \psr{92.80} & \va{52.12} & \, \tar{0.70} & \psr{94.13} & \va{52.28} & \, \tar{0.70} & \psr{93.10} & \va{52.92} & \, \tar{0.57} & \psr{90.29}\\
    $k$-Same-Select ($k$=10) & Class. & \va{57.20} & \, \tar{0.40} & \psr{95.57} & \va{58.40} & \, \tar{0.60} & \psr{94.63} & \va{57.72} & \, \tar{0.30} & \psr{93.30} & \va{52.10} & \, \tar{0.33} & \psr{94.86} & \va{52.95} & \, \tar{0.43} & \psr{93.96} & \va{51.53} & \, \tar{0.40} & \psr{91.53} \\
    $k$-Same-Furthest ($k$=10) & Class. & \va{56.80} & \, \tar{0.33} & \psr{95.30} & \va{58.03} & \, \tar{0.33} & \psr{94.37} & \va{57.09} & \, \tar{0.23} & \psr{93.33} & \va{52.10} & \, \tar{0.20} & \psr{95.10} & \va{52.57} & \, \tar{0.20} & \psr{93.76} & \va{52.07} & \, \tar{0.17} & \psr{91.16} \\
    \midrule
    MI-FGSM~\cite{dong2018boosting} ($\epsilon$=8) & Adv. & \va{95.77} & \tar{80.13} & \psr{16.47} & \va{98.09} & \tar{94.17} & \, \psr{5.80} & \va{98.71} & \tar{96.47} & \, \psr{1.90} & \va{94.53} & \tar{82.52} & \psr{12.55} & \va{95.38} & \tar{85.42} & \, \psr{6.87} & \va{95.53} & \tar{82.12} & \, \psr{4.47} \\ 
    PGD~\cite{madry2018towards} ($\epsilon$=8) & Adv. & \va{97.29} & \tar{91.93} & \, \psr{7.70} & \va{98.85} & \tar{97.03} & \, \psr{2.07} & \va{99.02} & \tar{97.97} & \, \psr{0.83} & \va{95.63} & \tar{88.76} & \, \psr{6.61} & \va{96.36} & \tar{88.39} & \, \psr{4.24} & \va{96.35} & \tar{85.79} & \, \psr{2.90} \\ 
    TI-DIM~\cite{dong2019evading} ($\epsilon$=8) & Adv. & \va{94.80} & \tar{79.23} & \psr{23.00} & \va{97.02} & \tar{90.80} & \psr{12.37} & \va{97.88} & \tar{93.63} & \, \psr{5.10} & \va{91.74} & \tar{67.99} & \psr{25.90} & \va{93.48} & \tar{69.33} & \psr{17.39} & \va{93.71} & \tar{77.24} & \psr{10.05} \\ 
    TIP-IM~\cite{yang2021towards} ($\epsilon$=8) & Adv. & \va{94.69} & \tar{79.10} & \psr{22.77} & \va{97.22} & \tar{86.90} & \psr{12.80} & \va{97.82} & \tar{93.33} & \, \psr{4.67} & \va{91.27} & \tar{61.58} & \psr{24.53} & \va{92.98} & \tar{70.56} & \psr{16.86} & \va{93.64} & \tar{75.90} & \, \psr{9.65} \\ 
    Chameleon~\cite{chow2024personalized} ($\epsilon$=8) & Adv. & \va{97.68} & \tar{91.53} & \, \psr{6.63} & \va{99.06} & \tar{97.97} & \, \psr{1.53} & \va{99.18} & \tar{98.10} & \, \psr{0.40} & \va{96.20} & \tar{89.89} & \, \psr{5.94} & \va{96.68} & \tar{88.73} & \, \psr{3.30} & \va{96.62} & \tar{89.26} & \, \psr{2.13} \\ 
    \midrule
    Adv-Makeup~\cite{yin2021adv} & Gen. & \va{99.25} & \tar{98.37} & \, \psr{0.00} & \va{99.23} & \tar{98.33} & \, \psr{0.00} & \va{99.23} & \tar{98.37} & \, \psr{0.03} & \va{97.98} & \tar{95.23} & \, \psr{0.27} & \va{97.93} & \tar{94.76} & \, \psr{0.10} & \va{97.57} & \tar{93.53} & \, \psr{0.27} \\
    CIAGAN~\cite{maximov2020ciagan} & Gen. & \va{74.82} & \, \tar{7.33} & \psr{59.83} & \va{75.85} & \, \tar{6.40} & \psr{52.20} & \va{73.18} & \, \tar{7.73} & \psr{46.93} & \va{60.95} & \, \tar{2.27} & \psr{77.08} & \va{61.79} & \, \tar{2.10} & \psr{68.25} & \va{62.35} & \, \tar{2.03} & \psr{55.34} \\
    AMT-GAN~\cite{hu2022protecting} & Gen. & \va{96.29} & \tar{82.70} & \, \psr{3.90} & \va{95.85} & \tar{77.77} & \, \psr{2.50} & \va{95.15} & \tar{70.80} & \, \psr{2.13} & \va{93.53} & \tar{60.74} & \, \psr{5.97} & \va{92.01} & \tar{54.17} & \, \psr{4.14} & \va{90.81} & \tar{46.80} & \, \psr{3.97} \\
    DeID-rPPG~\cite{savic2023identification} & Gen. & \va{99.25} & \tar{98.37} & \, \psr{0.00} & \va{99.23} & \tar{98.37} & \, \psr{0.00} & \va{99.22} & \tar{98.37} & \, \psr{0.03} & \va{97.92} & \tar{95.10} & \, \psr{0.57} & \va{97.75} & \tar{95.00} & \, \psr{0.57} & \va{97.45} & \tar{93.66} & \, \psr{0.43} \\
    G$^{2}$Face~\cite{yang2024g} & Gen. & \va{96.94} & \tar{87.00} & \, \psr{7.47} & \va{96.98} & \tar{87.90} & \, \psr{8.23} & \va{95.91} & \tar{75.70} & \, \psr{8.90} & \va{92.51} & \tar{62.98} & \psr{20.71} & \va{91.86} & \tar{55.57} & \psr{19.65} & \va{89.95} & \tar{52.77} & \psr{18.41} \\
    WeakenDiff~\cite{salar2025enhancing} & Gen. & \va{91.05} & \tar{43.17} & \psr{21.13} & \va{92.51} & \tar{54.63} & \psr{12.73} & \va{90.86} & \tar{56.17} & \psr{13.47} & \va{90.52} & \tar{54.66} & \psr{10.33} & \va{90.71} & \tar{39.28} & \, \psr{6.37} & \va{89.02} & \tar{38.23} & \, \psr{6.33}\\
    \bottomrule
    \end{tabular}
    }
    \vspace{-6mm}
\end{table}

\subsection{Benchmark Results}

We conduct comprehensive benchmarking of 17 FDeID methods re-implemented within \toolbox, evaluating privacy protection (\Cref{tab:privacy}), utility preservation (\Cref{tab:utility}), and visual quality (\Cref{tab:quality}). All methods are evaluated under identical conditions using our standardized pipelines and all reported numbers are produced by our own implementations.
We do not report numbers taken directly from original papers, as differences in evaluation protocols, datasets, and recognition models across studies make such numbers incomparable.

\noindent\textbf{Privacy Protection.} 
\Cref{tab:privacy} reveals substantial performance disparities across method paradigms. Classical $k$-anonymity methods, particularly $k$-Same-Select and $k$-Same-Furthest, achieve the strongest privacy protection with PSR exceeding 90\% across all recognition models and datasets, while maintaining verification accuracy near chance level ($\sim$50-58\%). These methods effectively disrupt identity features by replacing faces with carefully selected alternatives from the $k$-neighbor set. In contrast, adversarial perturbation methods demonstrate limited effectiveness under our evaluation protocol: despite their theoretical foundation in fooling recognition systems, methods like PGD and Chameleon achieve PSR below 10\% on LFW, indicating that the perturbations optimized against surrogate models (ArcFace) fail to transfer reliably to unseen recognizers. Among generative methods, CIAGAN provides moderate privacy protection (PSR of 46-77\% depending on the recognizer), while Adv-Makeup and DeID-rPPG show near-zero PSR, suggesting their modifications are insufficient to alter identity-discriminative features. Traditional naive methods exhibit inconsistent behavior: pixelation at $16\times16$ resolution achieves low verification accuracy but also low PSR due to complete facial structure destruction, whereas Gaussian blur preserves too much identity information even at $\sigma$=20. These findings highlight that effective privacy protection requires explicit identity feature manipulation rather than generic image degradation or imperceptible perturbations.

\begin{table}[t]
    \centering
    \caption{\textbf{Utility preservation performance across six downstream tasks.} We evaluate age estimation (MAE$\downarrow$) on AgeDB~\cite{moschoglou2017agedb}, gender and ethnicity classification (Acc$\uparrow$) on FairFace~\cite{karkkainen2021fairface}, expression recognition (Acc$\uparrow$) on AffectNet~\cite{mollahosseini2017affectnet}, landmark detection (NME$\downarrow$) on CelebA-HQ~\cite{karras2018progressive}, and rPPG estimation (HR-MAE$\downarrow$) on PURE~\cite{stricker2014non}. All results are from our unified re-implementation under identical conditions.}
    \label{tab:utility}
    \setlength{\tabcolsep}{4pt}
    \resizebox{0.95\textwidth}{!}{
    \begin{tabular}{l|c|cccccc}
    \toprule
    \multirow{2}{*}{\textbf{Method}} & \multirow{2}{*}{\textbf{Type}} & \textbf{Age} & \textbf{Gender} & \textbf{Ethnicity} & \textbf{Expression} & \textbf{Landmark} & \textbf{rPPG}  \\
    & & MAE$\downarrow$ & Acc$\uparrow$ & Acc$\uparrow$ & Acc$\uparrow$ & NME$\downarrow$ & HR-MAE$\downarrow$  \\
    \midrule
    \rowcolor{gray!10}
    Original & -- & 10.18 & 94.39 & 71.99 & 70.54 & 0.3601 & \, 0.39\\
    \midrule
    Gaussian Blur ($\sigma$=10) & Naive & \age{12.00} & \acc{93.97} & \acc{71.35} & \acc{54.09} & \nme{0.3609} & \, \rppg{0.40} \\
    Gaussian Blur ($\sigma$=20) & Naive & \age{14.18} & \acc{93.55} & \acc{69.33} & \acc{24.56} & \nme{0.3620} & \, \rppg{0.40} \\
    Pixelation (8$\times$8) & Naive & \age{15.26} & \acc{90.48} & \acc{59.76} & \acc{14.79} & \nme{0.3615} & \, \rppg{4.10} \\
    Pixelation (16$\times$16) & Naive & \age{18.93} & \acc{70.40} & \acc{17.12} & \acc{15.55} & \nme{0.3772} & \, \rppg{4.50} \\
    Black Mask & Naive & \age{18.31} & \acc{65.36} & \acc{20.26} & \acc{14.63} & \nme{0.8733} & \rppg{17.56} \\
    \midrule
    $k$-Same ($k$=5) & Class. & \age{15.91} & \acc{70.74} & \acc{27.30} & \acc{18.01} & \nme{0.3751} & \rppg{24.13} \\
    $k$-Same ($k$=10) & Class. & \age{16.00} & \acc{71.90} & \acc{27.52} & \acc{20.27} & \nme{0.3802} & \rppg{23.40} \\
    $k$-Same-Select ($k$=5) & Class. & \age{17.64} & \acc{60.22} & \acc{21.03} & \acc{15.87} & \nme{0.3923} & \rppg{31.11} \\
    $k$-Same-Select ($k$=10) & Class. & \age{17.81} & \acc{59.16} & \acc{20.49} & \acc{15.55} & \nme{0.3973} & \rppg{32.56} \\
    $k$-Same-Furthest ($k$=10) & Class. & \age{18.38} & \acc{56.43} & \acc{20.39} & \acc{15.35} & \nme{0.3974} & \rppg{28.37} \\
    \midrule
    MI-FGSM~\cite{dong2018boosting} ($\epsilon$=8) & Adv. & \age{10.29}  & \acc{91.93} & \acc{66.16} & \acc{69.26} & \nme{0.3601} & \, \rppg{1.70} \\
    PGD~\cite{madry2018towards} ($\epsilon$=8) & Adv. &  \age{10.04}  & \acc{93.31} & \acc{69.44} & \acc{69.91} & \nme{0.3599} & \, \rppg{1.65} \\
    TI-DIM~\cite{dong2019evading} ($\epsilon$=8) & Adv. & \age{10.91} & \acc{88.91} & \acc{61.05} & \acc{68.70} & \nme{0.3628} & \, \rppg{5.37} \\
    TIP-IM~\cite{yang2021towards} ($\epsilon$=8) & Adv. & \age{10.95} & \acc{88.64} & \acc{61.59} & \acc{68.90} & \nme{0.3630} & \, \rppg{3.60} \\
    Chameleon~\cite{chow2024personalized} ($\epsilon$=8) & Adv. & \, \age{9.95} & \acc{93.25} & \acc{70.19} & \acc{70.30} & \nme{0.3599} & \, \rppg{1.96}  \\
    \midrule
    Adv-Makeup~\cite{yin2021adv} & Gen. & \age{10.28} & \acc{94.05} & \acc{70.93} & \acc{70.48} & \nme{0.3613} & \, \rppg{0.40}  \\
    CIAGAN~\cite{maximov2020ciagan} & Gen. & \age{18.47} & \acc{56.56} & \acc{19.89} & \acc{28.99} & \nme{0.4242} & \, \rppg{9.01} \\
    AMT-GAN~\cite{hu2022protecting} & Gen. & \age{16.95} & \acc{92.46} & \acc{64.53} & \acc{66.28} & \nme{0.3595} & \rppg{26.17} \\
    DeID-rPPG~\cite{savic2023identification} & Gen. & \age{10.78} & \acc{94.03} & \acc{70.28} & \acc{69.38} & \nme{0.3627} & \, \rppg{0.39}  \\
    G$^{2}$Face~\cite{yang2024g} & Gen. & \age{11.82} & \acc{90.61} & \acc{63.61} & \acc{63.27} & \nme{0.3629} & \rppg{20.16} \\
    WeakenDiff~\cite{salar2025enhancing} & Gen. & \age{11.46} & \acc{86.63} & \acc{55.24} & \acc{65.36} & \nme{0.3603} & \rppg{13.90} \\
    \bottomrule
    \end{tabular}
    }
    \vspace{-7mm}
\end{table}

\noindent\textbf{Utility Preservation.} 
\Cref{tab:utility} demonstrates the inherent trade-off between privacy protection and utility preservation across FDeID methods. Adversarial perturbation methods excel at utility preservation, with Chameleon and PGD maintaining near-original performance across all six attributes (age MAE of 9.95-10.04 \textit{vs.}\ original 10.18; expression accuracy of 69.91-70.30\% \textit{vs.}\ original 70.54\%). This is expected since these methods introduce only minimal, imperceptible modifications to the input images. Generative methods exhibit mixed utility profiles: DeID-rPPG preserves rPPG signals exceptionally well (HR-MAE of 0.39, identical to original) and maintains strong demographic attribute accuracy, while Adv-Makeup achieves similar preservation due to its localized makeup-region modifications. However, CIAGAN suffers severe utility degradation across all attributes (gender accuracy drops to 56.56\%, expression to 28.99\%), reflecting its aggressive face replacement strategy. The $k$-Same family methods show consistently poor utility preservation, with $k$-Same-Furthest exhibiting the worst performance (age MAE of 18.38, gender accuracy of 56.43\%, rPPG HR-MAE of 28.37), as maximizing identity dissimilarity inherently disrupts attribute-correlated features. Naive methods demonstrate attribute-dependent degradation: Gaussian blur preserves demographic attributes reasonably well but severely impacts expression recognition (24.56\% at $\sigma$=20), while black masking catastrophically fails for landmark detection (NME of 0.8733) and rPPG estimation (HR-MAE of 17.56). These results underscore the importance of multi-attribute utility evaluation, as methods optimized for specific downstream tasks may fail unexpectedly on others.

\begin{table}[t]
    \centering
    \caption{\textbf{Visual quality assessment on LFW~\cite{huang2008labeled} and AgeDB~\cite{moschoglou2017agedb}.} PSNR$\uparrow$ and SSIM$\uparrow$ measure pixel-level fidelity; FID$\downarrow$ and NIQE$\downarrow$ assess perceptual realism; LPIPS$\downarrow$ quantifies perceptual distance. All results are from our unified re-implementation.}
    \label{tab:quality}
    \setlength{\tabcolsep}{3pt}
    \resizebox{1.0\textwidth}{!}{
    \begin{tabular}{l|c|ccccc|ccccc}
    \toprule
    \multirow{3}{*}{\textbf{Method}} & \multirow{3}{*}{\textbf{Type}} & \multicolumn{5}{c|}{\textbf{LFW}} & \multicolumn{5}{c}{\textbf{AgeDB}} \\
    \cmidrule(lr){3-7} \cmidrule(lr){8-12}
    &   & \textbf{PSNR$\uparrow$} & \textbf{SSIM$\uparrow$} & \textbf{FID$\downarrow$} & \textbf{NIQE$\downarrow$} &  \textbf{LPIPS$\downarrow$} & \textbf{PSNR$\uparrow$} & \textbf{SSIM$\uparrow$} & \textbf{FID$\downarrow$} & \textbf{NIQE$\downarrow$} &  \textbf{LPIPS$\downarrow$} \\
    \midrule
    Gaussian Blur ($\sigma$=10) & Naive & \psnr{35.90} & \ssim{0.9657} & \, \, \fid{4.28} & \niqe{7.7854} & \lpips{0.0470} & \psnr{31.02} & \ssim{0.8957} & \, \fid{12.81} & \, \niqe{9.5337} & \lpips{0.1515}  \\
    Gaussian Blur ($\sigma$=20) & Naive & \psnr{31.89} & \ssim{0.9339} & \, \fid{11.81} & \niqe{8.4497} & \lpips{0.1033} & \psnr{27.82} & \ssim{0.8353} & \, \fid{31.82} & \niqe{11.1360} & \lpips{0.2447}  \\
    Pixelation (8$\times$8) & Naive & \psnr{28.13} & \ssim{0.8862} & \, \fid{43.43} & \niqe{6.4532} & \lpips{0.2091} & \psnr{24.22} & \ssim{0.7441} & \fid{101.97} & \, \niqe{7.1216} & \lpips{0.3970}  \\
    Pixelation (16$\times$16) & Naive & \psnr{24.52} & \ssim{0.8555} & \, \fid{55.47} & \niqe{7.5537} & \lpips{0.2193} & \psnr{20.90} & \ssim{0.6846} & \fid{133.82} & \, \niqe{9.5578} & \lpips{0.4159}  \\
    Black Mask & Naive & \psnr{12.64} & \ssim{0.7500} & \fid{101.92} & \niqe{9.3674} & \lpips{0.2474} & \, \psnr{8.67} & \ssim{0.4542} & \fid{179.04} & \niqe{16.0312} & \lpips{0.4637}\\
    \midrule
    $k$-Same ($k$=5) & Class. & \psnr{26.15} & \ssim{0.9000} & \, \, \fid{7.19} & \niqe{5.6633} & \lpips{0.0878} & \psnr{21.20} & \ssim{0.7742} & \, \fid{27.46} & \, \niqe{5.3888} & \lpips{0.2162} \\
    $k$-Same ($k$=10) & Class. & \psnr{26.28} & \ssim{0.9029} & \, \, \fid{9.44} & \niqe{6.0672} & \lpips{0.0884} & \psnr{21.34} & \ssim{0.7815} & \, \fid{32.33} & \, \niqe{5.8310} & \lpips{0.2175} \\
    $k$-Same-Select ($k$=5) & Class. & \psnr{24.26} & \ssim{0.8823} & \, \, \fid{4.97} & \niqe{5.2572} & \lpips{0.0988} & \psnr{19.23} & \ssim{0.7382} & \, \fid{14.42} & \, \niqe{5.3431} & \lpips{0.2334} \\
    $k$-Same-Select ($k$=10) & Class. & \psnr{24.04} & \ssim{0.8808} & \, \, \fid{4.96} & \niqe{5.2356} & \lpips{0.1000} & \psnr{18.99} & \ssim{0.7350} & \, \fid{14.77} & \, \niqe{5.3041} & \lpips{0.2366} \\
    $k$-Same-Furthest ($k$=10) & Class. & \psnr{23.72} & \ssim{0.8786} & \, \, \fid{5.06} & \niqe{5.1978} & \lpips{0.1015} & \psnr{18.63} & \ssim{0.7304} & \, \fid{15.53} & \, \niqe{5.2617} & \lpips{0.2411} \\
    \midrule
    MI-FGSM~\cite{dong2018boosting} ($\epsilon$=8) & Adv. & \psnr{37.78} & \ssim{0.9561} & \, \fid{12.22} & \niqe{4.5203} & \lpips{0.0768} & \psnr{34.81} & \ssim{0.9165} & \, \fid{21.25} & \, \niqe{4.3010} & \lpips{0.1503} \\
    PGD~\cite{madry2018towards} ($\epsilon$=8) & Adv. & \psnr{40.21} & \ssim{0.9720} & \, \, \fid{5.98} & \niqe{4.6529} & \lpips{0.0529} & \psnr{37.16} & \ssim{0.9466} & \, \fid{11.86} & \, \niqe{4.4361} & \lpips{0.1100}  \\
    TI-DIM~\cite{dong2019evading} ($\epsilon$=8) & Adv. & \psnr{37.65} & \ssim{0.9744} & \, \, \fid{5.32} & \niqe{5.0848} & \lpips{0.0500} & \psnr{34.32} & \ssim{0.9517} & \, \fid{18.64} & \, \niqe{4.7763} & \lpips{0.1259}  \\
    TIP-IM~\cite{yang2021towards} ($\epsilon$=8) & Adv. & \psnr{37.68} & \ssim{0.9744} & \, \, \fid{5.39} & \niqe{5.0823} & \lpips{0.0500} & \psnr{34.35} & \ssim{0.9517} & \, \fid{18.61} & \, \niqe{4.7778} & \lpips{0.1256} \\
    Chameleon~\cite{chow2024personalized} ($\epsilon$=8) & Adv. & \psnr{41.28} & \ssim{0.9809} & \, \, \fid{4.40} & \niqe{4.6381} & \lpips{0.0418} & \psnr{38.21} & \ssim{0.9656} & \, \fid{11.59} & \, \niqe{4.6091} & \lpips{0.0895} \\
    \midrule
    Adv-Makeup~\cite{yin2021adv} & Gen. & \psnr{46.53} & \ssim{0.9961} & \, \, \fid{0.13} & \niqe{5.5207} & \lpips{0.0049} & \psnr{41.49} & \ssim{0.9937} & \, \, \fid{1.60} & \, \niqe{5.2059} & \lpips{0.0120} \\
    CIAGAN~\cite{maximov2020ciagan} & Gen. & \psnr{25.71} & \ssim{0.9065} & \, \, \fid{4.12} & \niqe{6.9326} & \lpips{0.0825} & \psnr{19.58} & \ssim{0.7349} & \, \fid{43.84} & \, \niqe{8.1664} & \lpips{0.2884}  \\
    AMT-GAN~\cite{hu2022protecting} & Gen. & \psnr{25.73} & \ssim{0.9383} & \, \, \fid{5.62} & \niqe{5.2237} & \lpips{0.0837} & \psnr{25.89} & \ssim{0.9422} & \, \fid{12.73} & \, \niqe{5.2373} & \lpips{0.0837} \\
    DeID-rPPG~\cite{savic2023identification} & Gen. & \psnr{35.71} & \ssim{0.9893} & \, \, \fid{0.84} & \niqe{5.7132} & \lpips{0.0218} & \psnr{28.59} & \ssim{0.9478} & \, \, \fid{9.10} & \, \niqe{6.0071} & \lpips{0.1245}  \\
    G$^{2}$Face~\cite{yang2024g} & Gen. & \psnr{32.32} & \ssim{0.9676} & \, \, \fid{2.51} & \niqe{5.3526} & \lpips{0.0352} & \psnr{28.05} & \ssim{0.9252} & \, \fid{14.22} & \, \niqe{5.0341} & \lpips{0.1015} \\
    WeakenDiff~\cite{salar2025enhancing} & Gen. & \psnr{28.84} & \ssim{0.8700} & \, \, \fid{9.08} & \niqe{5.9011} & \lpips{0.1563} & \psnr{27.48} & \ssim{0.8179} & \, \, \fid{9.61} & \, \niqe{5.5688} & \lpips{0.2030} \\
    \bottomrule
    \end{tabular}
    }
    \vspace{-6mm}
\end{table}

\noindent\textbf{Visual Quality.} 
\Cref{tab:quality} reveals that different metrics capture complementary aspects of visual quality, and single-metric evaluation can be misleading.
Adv-Makeup achieves the highest pixel-level and perceptual fidelity across the board because it restricts modifications to the makeup region, leaving the remaining facial structure intact. DeID-rPPG similarly attains high structural fidelity (PSNR of 35.71\,dB, FID of 0.84) through its subtle temporal adjustments.
Among adversarial methods, Chameleon achieves the best balance (PSNR of 41.28\,dB, FID of 4.40, NIQE of 4.64), while MI-FGSM attains a comparable PSNR but incurs a much higher FID, indicating that its structured gradient-aligned noise introduces feature-level distributional shifts not captured by pixel metrics alone. A concrete example of inter-metric disagreement also seen in AMT-GAN, which achieves only moderate PSNR yet competitive FID because its synthesis produces perceptually realistic outputs.
A notable cross-dataset finding is that CIAGAN's FID deteriorates sharply from 4.12 on LFW to 43.84 on AgeDB, exposing distributional fragility beyond its training domain.
$k$-Same methods yield moderate fidelity but competitive NIQE scores, since they substitute real face images whose natural texture statistics are well-preserved; black masking, by contrast, yields the worst results across all metrics.
Counterintuitively, adversarial methods achieve the best NIQE scores (4.52--5.08) among all paradigms, outperforming even generative methods (5.22--6.93), because their perturbations are constrained to a tight $\ell_\infty$ ball that preserves low-level natural image statistics.
Collectively, these results confirm a general inverse relationship between privacy protection and visual quality, and underscore the necessity of multi-metric evaluation for reliable quality assessment in FDeID.

\subsection{Privacy-Utility-Quality Trade-off Analysis}

Jointly analyzing \Cref{tab:privacy}--\Cref{tab:quality}, methods cluster into distinct operating regimes.
At one extreme, adversarial methods (PGD, Chameleon) and localized generative methods (Adv-Makeup, DeID-rPPG) achieve near-original utility and excellent visual quality but provide negligible privacy protection (PSR${<}$4\%), as perturbations optimized against surrogate recognizers fail to transfer reliably to unseen models.
At the opposite extreme, $k$-Same-Select and $k$-Same-Furthest deliver robust privacy (PSR${>}$93\%) at the cost of severe utility degradation (gender accuracy falling to 56--60\%, rPPG HR-MAE deteriorating from 0.39 to ${>}$28) and moderate quality reduction (PSNR of 23--24\,dB).
Between these extremes, CIAGAN achieves moderate privacy but occupies an unfavorable position: it simultaneously incurs severe utility loss (gender accuracy of 56.56\%, expression of 28.99\%) and cross-domain quality fragility (FID rising from 4.12 on LFW to 43.84 on AgeDB), offering no compelling advantage over either regime.
More nuanced attribute-level trade-offs emerge in other methods: G$^{2}$Face achieves good visual quality (PSNR of 32.32\,dB, FID of 2.51) yet its rPPG utility degrades substantially (HR-MAE of 20.16), showing that strong appearance fidelity does not guarantee preservation of physiological signals; AMT-GAN preserves facial landmarks best of all methods (NME of 0.3595) and maintains high gender accuracy (92.46\%), yet exhibits the worst rPPG among generative methods (HR-MAE of 26.17), underscoring that utility preservation is highly attribute-specific.

These observations reveal two patterns: utility and visual quality are \textit{positively correlated}, since both benefit from minimal image modification, while both exhibit \textit{negative correlations} with privacy protection effectiveness.
Crucially, no evaluated method simultaneously achieves strong privacy (PSR${>}$90\%), high utility ($<$2\% relative degradation across all attributes), and excellent visual quality (PSNR${>}$35\,dB), confirming an open research challenge: developing FDeID techniques that can selectively suppress identity-discriminative features while preserving the full set of attribute-relevant and appearance information.
\toolbox\ provides the standardized infrastructure necessary to track progress and benchmark future methods under reproducible, consistent conditions.

\section{Additional Features}

\subsection{Method Ensemble}

A unique capability enabled by our unified interface is \textit{method ensemble}, which systematically combines multiple FDeID methods to achieve privacy-utility-quality trade-offs unattainable by any single method. We explore three ensemble strategies, with results summarized in \Cref{tab:ensemble}.

\noindent\textbf{Sequential Ensemble.} Methods are applied in a pipeline, where the output of one method becomes the input to the next. This is particularly effective for combining complementary approaches, \eg, a generative method that alters facial structure followed by an adversarial method that adds perturbations to confuse residual identity features:
\begin{equation}
    \mathbf{x}_{\text{deid}} = f_n \circ f_{n-1} \circ \cdots \circ f_1(\mathbf{x}_{\text{orig}}),
\end{equation}
where $f_i$ denotes the $i$-th FDeID method. As shown in \Cref{tab:ensemble}, CIAGAN$\rightarrow$TI-DIM boosts privacy protection, raising PSR from 53.0\% (CIAGAN alone) to 77.8\% while reducing TAR from 7.15\% to 3.00\%. Meanwhile, $k$-Same-Select$\rightarrow$TI-DIM improves utility preservation (\eg, gender accuracy increases from 60.2\% to 87.8\%) at the cost of weaker privacy compared to $k$-Same-Select alone.

\begin{table}[t]
    \centering
    \caption{\textbf{Method ensemble results.} Different ensemble strategies achieve varied privacy-utility-quality trade-offs. VA, TAR, and PSR are averaged across three recognizers (ArcFace~\cite{deng2019arcface}, CosFace~\cite{wang2018cosface}, AdaFace~\cite{kim2022adaface}). Best results are highlighted in bold.}
    \label{tab:ensemble}
    \setlength{\tabcolsep}{3pt}
    \resizebox{\textwidth}{!}{
    \begin{tabular}{l|l|ccc|cccc|c}
    \toprule
    \multirow{2}{*}{\textbf{Ensemble Type}} & \multirow{2}{*}{\textbf{Configuration}} & \multicolumn{3}{c|}{\textbf{Privacy}} & \multicolumn{4}{c|}{\textbf{Utility}} & \textbf{Quality} \\
    \cmidrule(lr){3-5} \cmidrule(lr){6-9} \cmidrule(lr){10-10}
    & & VA$\downarrow$ & TAR$\downarrow$ & PSR$\uparrow$ & Age$\downarrow$ & Gender$\uparrow$ & Expr$\uparrow$ & rPPG$\downarrow$ & PSNR$\uparrow$ \\
    \midrule
    \rowcolor{gray!10}
    \multicolumn{10}{l}{\textit{Single Methods (Baselines)}} \\
    -- & CIAGAN & 74.62 & \, 7.15 & 52.99 & 18.47 & 56.56 & 28.99 & \, 9.01 & 25.71 \\
    -- & TI-DIM ($\epsilon$=8) & 96.57 & 87.89 & 13.49 & 10.91 & 88.91 & 68.70 & \, 5.37 & 37.65 \\
    -- & $k$-Same-Select ($k$=5) & \textbf{57.85} & \, \textbf{0.76} & \textbf{93.75} & 17.64 & 60.22 & 15.87 & 31.11 & 24.26 \\
    -- & DeID-rPPG & 99.23 & 98.37 & \, 0.01 & 10.78 & \textbf{94.03} & 69.38 & \, \textbf{0.39} & 35.71 \\
    \midrule
    \rowcolor{blue!5}
    \multicolumn{10}{l}{\textit{Sequential Ensemble}} \\
    Sequential & CIAGAN $\rightarrow$ TI-DIM & 68.46 & \, 3.00 & 77.75 & 12.57 & 56.74 & 27.62 & 22.91 & 25.02 \\
    Sequential & $k$-Same-Select $\rightarrow$ TI-DIM & 86.96 & 43.54 & 46.99 & 12.57 & 87.79 & 36.57 & \, 4.07 & 33.69 \\
    Sequential & DeID-rPPG $\rightarrow$ CIAGAN & 73.69 & \, 5.18 & 54.94 & 18.95 & 54.90 & 28.76 & \, 7.55 & 24.34 \\
    \midrule
    \rowcolor{green!5}
    \multicolumn{10}{l}{\textit{Parallel Ensemble (Weighted Fusion)}} \\
    Parallel & 0.7$\times$CIAGAN + 0.3$\times$$k$-Same-Select & 77.70 & \, 8.72 & 46.42 & 17.32 & 65.62 & 31.77 & \, 8.52 & 27.77 \\
    Parallel & 0.5$\times$CIAGAN + 0.5$\times$DeID-rPPG & 96.17 & 85.38 & \, 8.67 & 13.14 & 82.75 & 59.71 & \, 7.01 & 30.85 \\
    Parallel & 0.6$\times$$k$-Same-Select + 0.4$\times$TI-DIM & 98.99 & 97.82 & \, 0.71 & 11.09 & 93.49 & 68.29 & \, 1.59 & \textbf{38.82} \\
    \midrule
    \rowcolor{red!5}
    \multicolumn{10}{l}{\textit{Attribute-Guided Ensemble}} \\
    Attr-Guided & preserve=\{gender, expr\}, suppress=\{identity\} & 96.99 & 89.85 & 10.42 & \textbf{10.75} & 89.47 & 69.20 & \, 2.38 & 38.17 \\
    Attr-Guided & preserve=\{age, rPPG\}, suppress=\{identity, ethnicity\} & 97.04 & 90.27 & 10.43 & 10.76 & 89.47 & \textbf{69.42} & \, 4.48 & 38.18 \\
    Attr-Guided & preserve=\{landmark\}, suppress=\{identity, gender\} & 96.96 & 89.41 & 10.84 & \textbf{10.75} & 89.37 & 69.20 & \, 3.61 & 38.17 \\
    \bottomrule
    \end{tabular}
    }
    \vspace{-6mm}
\end{table}

\noindent\textbf{Parallel Ensemble with Fusion.} 
Multiple methods process the input independently, and their outputs are combined via weighted blending:
\begin{equation}
    \mathbf{x}_{\text{deid}} = \sum_{i=1}^{n} w_i \cdot f_i(\mathbf{x}_{\text{orig}}), \quad \text{where} \sum_{i=1}^{n} w_i = 1.
\end{equation}
This strategy enables smooth interpolation between methods with complementary characteristics. Notably, 0.6$\times$$k$-Same-Select + 0.4$\times$TI-DIM achieves the highest PSNR of 38.82\,dB among all ensemble configurations while also attaining strong utility (gender: 93.5\%, expression: 68.3\%), though at the expense of privacy (TAR rises to 97.8\%). Conversely, 0.7$\times$CIAGAN + 0.3$\times$$k$-Same-Select maintains moderate privacy (TAR: 8.72\%) with improved utility over either constituent method alone.

\noindent\textbf{Attribute-Guided Ensemble.} 
Rather than manually selecting methods and weights, users specify which attributes to preserve or suppress, and the toolbox automatically configures an optimal ensemble based on per-method attribute preservation profiles from our benchmark.
As shown in \Cref{tab:ensemble}, all three attribute-guided configurations achieve consistently strong performance across metrics, with the \texttt{preserve=\{age, rPPG\}} variant obtaining the best expression accuracy (69.4\%) while maintaining high privacy (TAR: 90.3\%). Compared to manual parallel fusion, attribute-guided ensembles deliver more balanced trade-offs without requiring domain expertise to tune individual method weights, substantially lowering the barrier to deploying effective de-identification pipelines.

\subsection{Video Processing Pipeline}

Beyond image-based evaluation, \textbf{FDeID-Toolbox} provides a video processing pipeline that supports both video files and frame-sequence directories through a unified input abstraction. The pipeline follows a detect-then-deidentify paradigm: RetinaFace~\cite{deng2020retinaface} detects faces in each frame, and a selected method is applied to the detected face regions before reinsertion into the original frame. To enable real-time processing, we introduce a detection skipping strategy, which reuses bounding boxes from prior detections on intermediate frames, trading marginal spatial accuracy for substantial speedup.

\begin{figure}[tb]
  \centering
  \begin{subfigure}{0.42\linewidth}
    \includegraphics[width=1.0\columnwidth]{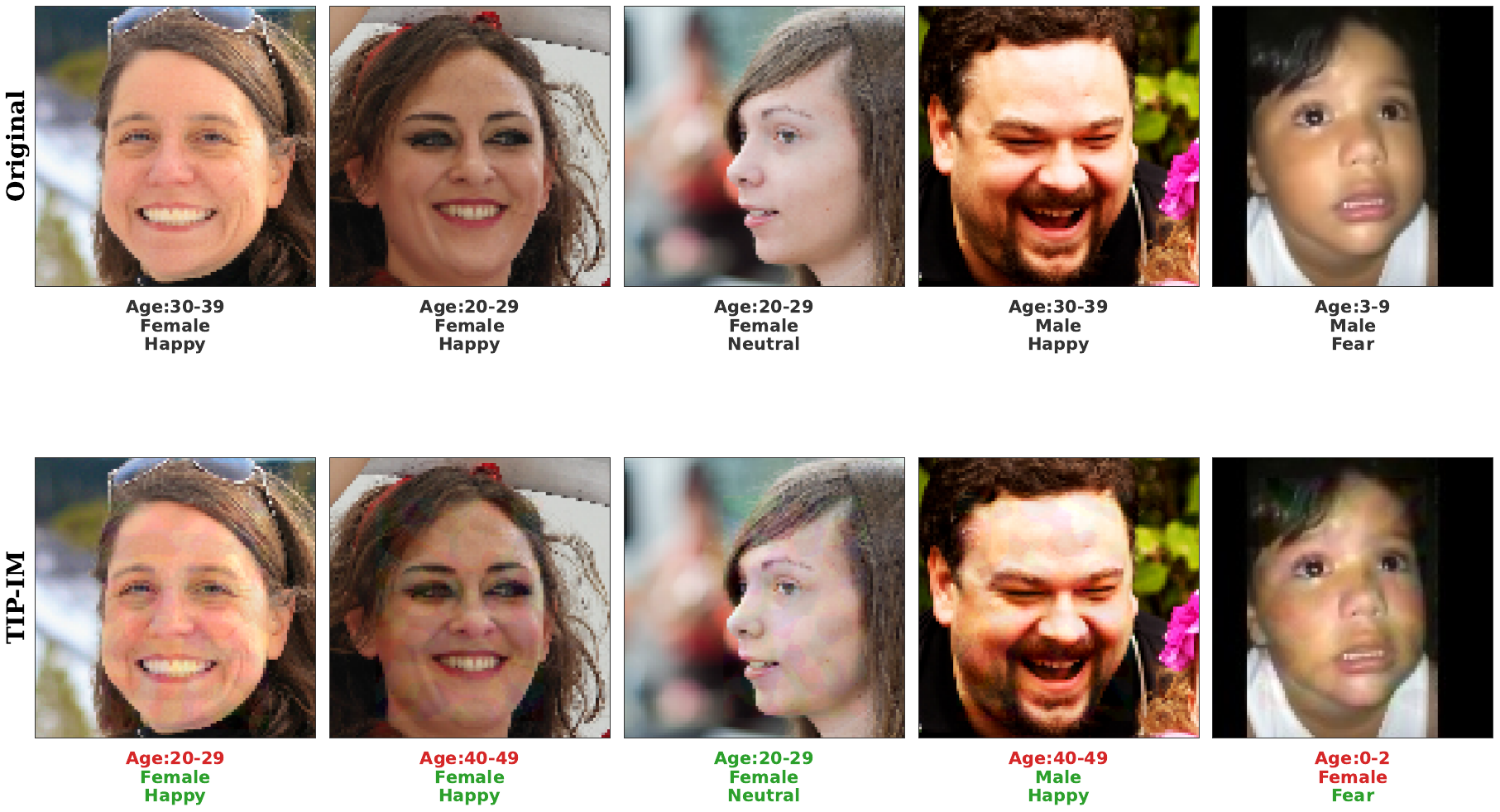}
    \caption{Attribute visualization}
    \label{fig:vis-a}
  \end{subfigure}
  \hfill
  \begin{subfigure}{0.31\linewidth}
    \includegraphics[width=1.0\columnwidth]{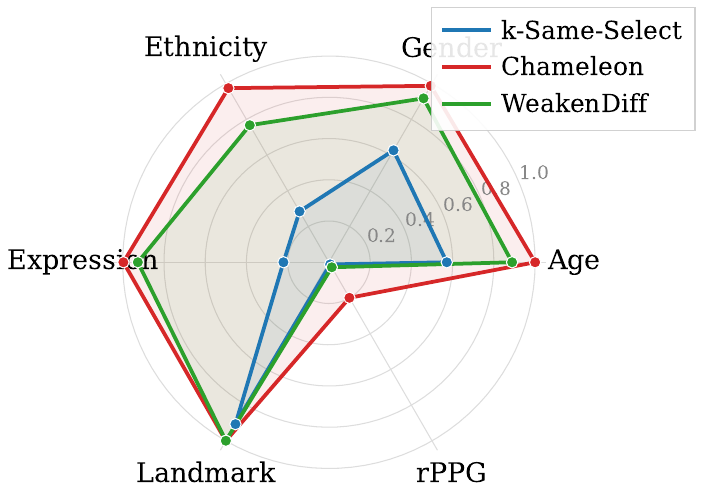}
    \caption{Radar chart}
    \label{fig:vis-b}
  \end{subfigure}
  \hfill
  \begin{subfigure}{0.25\linewidth}
    \includegraphics[width=1.0\columnwidth]{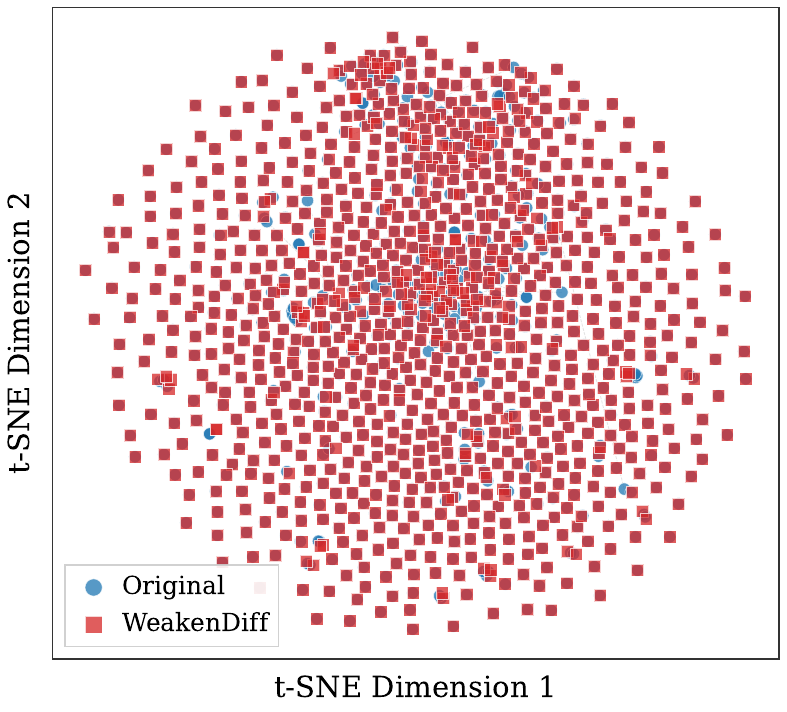}
    \caption{t-SEN visualization}
    \label{fig:vis-c}
  \end{subfigure}
  \vspace{-6mm}
  \caption{Examples of the \toolbox's visualization module.}
  \label{fig:visualization}
  \vspace{-6mm}
\end{figure}

\vspace{-2mm}
\subsection{Visualization and Analysis Tools}

To support qualitative analysis, we provide a suite of visualization functions (as shown in \Cref{fig:visualization}): (1) \textit{side-by-side comparison} generates grid visualizations comparing multiple methods on the same input; (2) \textit{attribute overlay} visualizes predicted attributes (age, gender, expression) on both original and de-identified faces to diagnose utility degradation; (3) \textit{metric radar charts} provide intuitive multi-dimensional comparison of methods across utilities; and (4) \textit{identity embedding visualization} projects face embeddings into 2D space using t-SNE~\cite{maaten2008visualizing} to show identity separation between original and de-identified faces.

\section{Conclusion}
We present \toolbox, a comprehensive and open-source for FDeID that addresses the longstanding challenges of fragmented implementations, inconsistent evaluation protocols, and incomparable results across studies. Our toolbox integrates 17 representative methods spanning four paradigms within a unified framework comprising modular data, method, pipeline, and evaluation components. Through benchmarking on mainstream datasets, we provide the systematic comparison across three evaluation dimensions: privacy protection, utility preservation, and visual quality. Our experiments reveal that no single method dominates all dimensions. We will release the codebase to facilitate reproducible research and accelerate progress in privacy-preserving computer vision.

\noindent\textbf{Limitations.} 
Despite the breadth of our toolbox, several limitations remain. First, our current evaluation is restricted to static images and does not address temporal consistency or identity leakage across video frames. Second, the current fairness analysis across demographic subgroups is limited, and a more thorough investigation of performance disparities with respect to age, gender, ethnicity, and skin tone is necessary to ensure equitable deployment of FDeID technologies.

\noindent\textbf{Broader Impacts.} 
FDeID is inherently dual-use, which can be misused to evade accountability or obstruct lawful identification. We encourage responsible use in compliance with applicable data protection regulations such as GDPR and urge practitioners to consider the ethical implications of their deployment contexts.

\bibliography{reference}
\bibliographystyle{splncs}

\appendix
\newpage
\input{supplementary}

\end{document}

%% file: math_commands.tex
\renewcommand{\Sigma}{\mathfrak{S}}

\def\eqref#1{equation~\ref{#1}}

\def\1{\bm{1}}

\newcommand{\test}{\mathcal{D_{\mathrm{test}}}}

\DeclareMathAlphabet{\mathsfit}{\encodingdefault}{\sfdefault}{m}{sl}
\SetMathAlphabet{\mathsfit}{bold}{\encodingdefault}{\sfdefault}{bx}{n}

%% file: supplementary.tex
\section{Outline}
\label{sec:outline}

This supplementary material provides more details of all face de-identification (FDeID) algorithms implemented in the \toolbox.
\Cref{sec:qualitative} shows additional qualitative results.
\Cref{sec:architecture} describes the overall design and the pretrained dependency models.
\Cref{sec:naive} details the three naive FDeID methods: Gaussian blur, pixelation, and masking.
\Cref{sec:ksame} presents the $k$-same family: $k$-Same-Average~\cite{newton2005preserving}, $k$-Same-Select~\cite{gross2005integrating}, and $k$-Same-Furthest~\cite{meng2014face}.
\Cref{sec:adversarial} describes the adversarial perturbation methods: PGD~\cite{madry2018towards}, MI-FGSM~\cite{dong2018boosting}, TI-DIM~\cite{dong2019evading}, TIP-IM~\cite{yang2021towards}, and Chameleon~\cite{chow2024personalized}.
\Cref{sec:generative} covers the generative methods: CIAGAN~\cite{maximov2020ciagan}, AMT-GAN~\cite{hu2022protecting}, Adv-Makeup~\cite{yin2021adv}, WeakenDiff~\cite{salar2025enhancing}, DeID-rPPG~\cite{savic2023identification}, and G$^{2}$Face~\cite{yang2024g}.

\section{Additional Qualitative Results}
\label{sec:qualitative}

\begin{figure}[t]
    \centering\small
    \includegraphics[width=1.0\columnwidth]{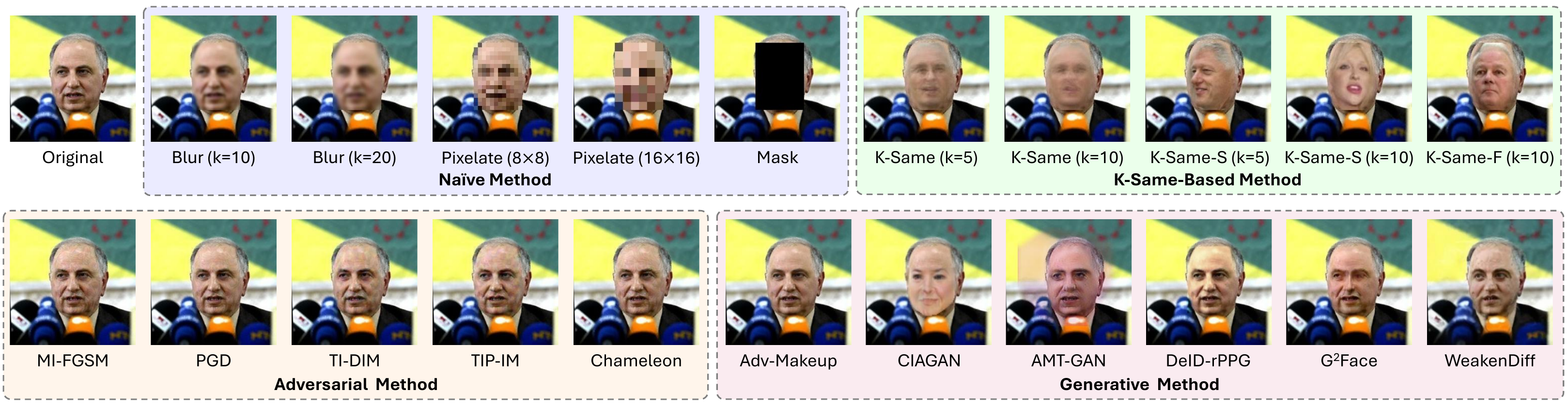}
    \vspace{-6mm}
    \caption{\textbf{Qualitative comparison of FDeID methods on a sample face image}. The source image is from LFW dataset~\cite{huang2008labeled}. Methods are grouped by paradigm: naive, $k$-same-based, adversarial, and generative.}
    \label{fig:show_supp1}
\end{figure}

\begin{figure}[t]
    \centering\small
    \includegraphics[width=1.0\columnwidth]{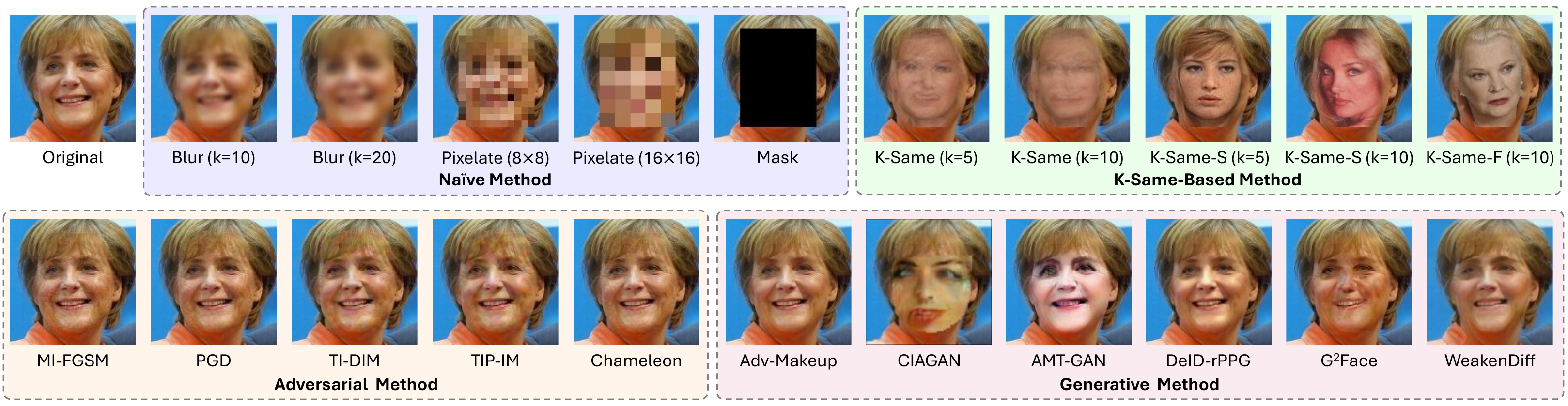}
    \vspace{-6mm}
    \caption{\textbf{Qualitative comparison of FDeID methods on a sample face image}. The source image is from AgeDB dataset~\cite{moschoglou2017agedb}.}
    \label{fig:show_supp2}
\end{figure}

\Cref{fig:show_supp1,fig:show_supp2} present comparisons of all 17 FDeID methods implemented in the \toolbox\ across two face images drawn from the LFW~\cite{huang2008labeled} and AgeDB~\cite{moschoglou2017agedb} dataset, respectively.
Each image is organized by paradigm: naive, $k$-same-based, adversarial, and generative.

\section{\toolbox\ Details}
\label{sec:architecture}

\subsection{Codebase Overview}

\begin{enumerate}[noitemsep]
    \item \textbf{Data Module} (\texttt{core/data/}): Unified data loaders and preprocessing pipelines for diverse face datasets with standardized annotation formats.
    \item \textbf{Method Module} (\texttt{core/fdeid/}): De-identification algorithms spanning naive, $k$-same, adversarial, and generative paradigms.
    \item \textbf{Ensemble Module} (\texttt{core/ensemble/}): Sequential, parallel, and attribute-guided method combination strategies.
    \item \textbf{Evaluation Module} (\texttt{core/eval/}, \texttt{core/identity/}, \texttt{core/utility/}): Privacy, utility, and quality metrics with pretrained model backends.
    \item \textbf{Visualization Module} (\texttt{core/visualization/}): Publication-quality figure generation including side-by-side comparisons, radar charts, t-SNE embedding plots, and attribute overlays.
\end{enumerate}

\subsection{Dependency Models}

The \toolbox\ relies on several pretrained models that serve as building blocks across the de-identification pipeline and evaluation framework.
These models can be categorized into three groups: face detection, face recognition, and utility attribute prediction.

\noindent\textbf{Face Detection.}
\textbf{RetinaFace}~\cite{deng2020retinaface} is used as the primary face detector throughout the toolbox.
It is based on a single-stage anchor-based detection framework and supports two backbone configurations: ResNet-50 and MobileNet-0.25.
RetinaFace outputs bounding boxes with confidence scores and 5-point facial landmarks (left eye, right eye, nose tip, left mouth corner, right mouth corner), which are subsequently used for face alignment.
Additionally, \textbf{Dlib}~\cite{kazemi2014one} is employed for 68-point facial landmark detection, which is required by several generative methods (\eg, CIAGAN, AMT-GAN, Adv-Makeup) for precise face alignment and region-specific processing.

\noindent\textbf{Face Recognition.}
Three face recognition models are integrated for both adversarial attack targets and privacy evaluation:
\begin{itemize}[noitemsep]
    \item \textbf{ArcFace}~\cite{deng2019arcface}: Uses a ResNet-100 backbone trained on MS1MV3 with additive angular margin loss. Produces 512-dimensional embeddings. It serves as the default surrogate model for all adversarial de-identification methods.
    \item \textbf{CosFace}~\cite{wang2018cosface}: Uses a ResNet-50 backbone trained on Glint360K with large-margin cosine loss. Produces 512-dimensional embeddings. It shares the same inference architecture as ArcFace but differs in the training loss formulation.
    \item \textbf{AdaFace}~\cite{kim2022adaface}: Employs an IR-SE backbone with adaptive margin based on image quality. Produces 512-dimensional embeddings. It includes built-in 5-point alignment utilities with a default crop size of $96 \times 112$.
\end{itemize}
All three models are used to compute cosine similarity scores between original and de-identified face embeddings during privacy evaluation.
The use of multiple recognition models provides a more robust privacy assessment, as different models may respond differently to de-identification artifacts.

\noindent\textbf{Utility Attribute Prediction.}
Several pretrained models are employed for evaluating the preservation of semantic face attributes after de-identification:
\begin{itemize}[noitemsep]
    \item \textbf{FairFace}~\cite{karkkainen2021fairface}: A ResNet-34 model trained for joint prediction of age (9 groups), gender (2 classes), and race/ethnicity (7 classes) with a unified 18-class output head. Input resolution: $224 \times 224$. It is used for evaluating age, gender, and ethnicity preservation in our toolbox.
    \item \textbf{POSTER}~\cite{zheng2023poster}: A pyramid cross-fusion transformer for facial expression recognition. It employs a dual-backbone architecture: MobileFaceNet ($112 \times 112$) for landmark-based geometric features and IR-50 ($224 \times 224$) for appearance features, fused through a vision transformer (embed\_dim=512, 8 heads). Classifies 7 basic expressions (neutral, happy, sad, surprise, fear, disgust, anger).
    \item \textbf{HRNet}~\cite{wang2020deep}: High-Resolution Network for facial landmark detection. Maintains multi-scale representations through four parallel branches with repeated multi-scale fusion. Input: $256 \times 256$; output: $64 \times 64$ heatmaps for 68 landmark points. The Normalized Mean Error (NME) between original and de-identified landmarks quantifies geometric preservation.
    \item \textbf{FactorizePhys}~\cite{joshi2024factorizephys}: A 3D CNN with factorized self-attention via non-negative matrix factorization for remote photoplethysmography (rPPG). Processes 160-frame video chunks at $72 \times 72$ resolution and extracts blood volume pulse (BVP) signals for non-contact heart rate estimation. Used to evaluate whether de-identification preserves the subtle color variations required for physiological signal extraction.
\end{itemize}

Table~\ref{tab:dependency_models} summarizes all dependency models with their key specifications.

\begin{table*}[t]
\centering
\caption{Summary of dependency models used in the \toolbox\ for face detection, recognition, and utility attribute evaluation.}
\label{tab:dependency_models}
\setlength{\tabcolsep}{4pt}
\resizebox{\textwidth}{!}{
\begin{tabular}{llllll}
\toprule
\textbf{Model} & \textbf{Task} & \textbf{Backbone} & \textbf{Input Size} & \textbf{Output}  \\
\midrule
RetinaFace~\cite{deng2020retinaface} & Face detection & ResNet-50 & Variable & Bbox + 5 landmarks  \\
Dlib~\cite{kazemi2014one} & Landmark detection & --- & Variable & 68 landmarks \\
\midrule
ArcFace~\cite{deng2019arcface} & Face recognition & ResNet-100 & $112 \times 112$ & 512-d embedding  \\
CosFace~\cite{wang2018cosface} & Face recognition & ResNet-50 & $112 \times 112$ & 512-d embedding  \\
AdaFace~\cite{kim2022adaface} & Face recognition & IR-SE & $112 \times 112$ & 512-d embedding  \\
\midrule
FairFace~\cite{karkkainen2021fairface} & Age/Gender/Ethnicity & ResNet-34 & $224 \times 224$ & 18 classes  \\
POSTER~\cite{zheng2023poster} & Expression & MobileFaceNet + IR-50 & $112$/$224$ & 7 classes + 512-d  \\
HRNet~\cite{wang2020deep} & Landmarks & HRNet-W18 & $256 \times 256$ & 68$\times$2 coords  \\
FactorizePhys~\cite{joshi2024factorizephys} & rPPG & 3D CNN + NMF-FSAM & $72 \times 72 \times 160$ & BVP signal  \\
\bottomrule
\end{tabular}
}
\vspace{-4mm}
\end{table*}

\section{Naive FDeID Methods}
\label{sec:naive}

Naive methods apply classical image processing transformations to obscure facial identity.
They are computationally efficient but generally offer limited utility preservation.
The toolbox implements three naive methods in \texttt{core/fdeid/naive/}.

\subsection{Gaussian Blur}

\textbf{Class:} \texttt{GaussianBlurDeIdentifier} \\
\textbf{Method name:} \texttt{blur}

Gaussian blur applies a low-pass filter to the face region.

\noindent\textbf{Parameters:}
\begin{itemize}[noitemsep]
    \item \texttt{kernel\_size} (int, default: 51): Gaussian kernel size (must be odd).
    \item \texttt{sigma} (float, default: 0): Gaussian sigma; 0 triggers automatic calculation based on kernel size.
\end{itemize}

\subsection{Pixelation}

\textbf{Class:} \texttt{PixelateDeIdentifier} \\
\textbf{Method name:} \texttt{pixelate}

Pixelation creates a mosaic effect by downsampling and upsampling the face region.

\noindent\textbf{Parameters:}
\begin{itemize}[noitemsep]
    \item \texttt{block\_size} (int, default: 16): Size of pixelation blocks.
    \item \texttt{interpolation} (str, default: \texttt{nearest}): Upsampling interpolation (\texttt{nearest} or \texttt{linear}).
\end{itemize}

\subsection{Mask}

\textbf{Class:} \texttt{MaskDeIdentifier} \\
\textbf{Method name:} \texttt{mask}

Replaces the entire face region with a solid color, providing strong privacy at the cost of complete attribute loss.

\noindent\textbf{Parameters:}
\begin{itemize}[noitemsep]
    \item \texttt{mask\_color} (tuple, default: $(0, 0, 0)$): RGB color for the mask.
    \item \texttt{mask\_type} (str, default: \texttt{solid}): One of \texttt{solid}, \texttt{random\_color}, \texttt{white}, \texttt{black}.
\end{itemize}

\section{$k$-Same FDeID Methods}
\label{sec:ksame}

The $k$-same family of methods~\cite{newton2005preserving} achieves $k$-anonymity by replacing each face with a synthetic face derived from $k$ similar faces in a reference dataset.
All $k$-same methods are implemented in \texttt{core/fdeid/ksame/methods.py} and use pixel-wise $L_2$ distance for similarity matching.

\subsection{$k$-Same-Average}

\textbf{Class:} \texttt{KSameAverage} \\
\textbf{Method name:} \texttt{average}~\cite{newton2005preserving}

\noindent\textbf{Parameters:}
\begin{itemize}[noitemsep]
    \item \texttt{k} (int, default: 10): Number of similar faces to average.
    \item \texttt{reference\_dataset} (str): Path to reference face images directory.
    \item \texttt{face\_detector} (object): Face detector for extracting face crops from reference images.
\end{itemize}

\subsection{$k$-Same-Select}

\textbf{Class:} \texttt{KSameSelect} \\
\textbf{Method name:} \texttt{select}~\cite{gross2005integrating}

\noindent\textbf{Parameters:}
\begin{itemize}[noitemsep]
    \item \texttt{k} (int, default: 10): Number of similar faces to consider.
    \item \texttt{selection\_mode} (str): Selection strategy---\texttt{closest}, \texttt{furthest}, or \texttt{random}.
    \item \texttt{reference\_dataset} (str): Path to reference dataset.
\end{itemize}

\subsection{$k$-Same-Furthest}

\textbf{Class:} \texttt{KSameFurthest} \\
\textbf{Method name:} \texttt{furthest}~\cite{meng2014face}

\noindent\textbf{Parameters:}
\begin{itemize}[noitemsep]
    \item \texttt{k} (int, default: 10): Number of similar candidates.
    \item \texttt{reference\_dataset} (str): Path to reference dataset.
\end{itemize}

\section{Adversarial FDeID Methods}
\label{sec:adversarial}

Adversarial methods generate perturbations that cause face recognition models to fail.
All adversarial methods are implemented in \texttt{core/fdeid/adversarial/} and share the following design principles:
\begin{itemize}[noitemsep]
    \item Perturbations are applied \emph{only} to the extracted face region.
    \item Optimization minimizes cosine similarity between original and perturbed face embeddings.
    \item A default ArcFace model (ResNet-100, 512-d embeddings) is automatically created if not provided.
\end{itemize}

\subsection{PGD}

\textbf{Class:} \texttt{PGDDeIdentifier} \\
\textbf{Method name:} \texttt{pgd}~\cite{madry2018towards} \\

\noindent\textbf{Parameters:}
\begin{itemize}[noitemsep]
    \item \texttt{epsilon} (float, default: $8/255$): Maximum perturbation budget.
    \item \texttt{alpha} (float, default: $2/255$): Step size per iteration.
    \item \texttt{num\_iter} (int, default: 20): Number of PGD iterations.
    \item \texttt{norm} (str, default: \texttt{Linf}): Norm constraint (\texttt{Linf}, \texttt{L2}, \texttt{L1}).
    \item \texttt{targeted} (bool, default: False): Targeted \emph{vs.}\ untargeted attack.
    \item \texttt{random\_start} (bool, default: True): Random initialization within $\epsilon$-ball.
\end{itemize}

\subsection{MI-FGSM}

\textbf{Class:} \texttt{MIFGSMDeIdentifier} \\
\textbf{Method name:} \texttt{mifgsm}~\cite{dong2018boosting} \\

\noindent\textbf{Parameters:}
\begin{itemize}[noitemsep]
    \item \texttt{epsilon} (float, default: $8/255$): Perturbation budget.
    \item \texttt{alpha} (float, default: $2/255$): Step size.
    \item \texttt{num\_iter} (int, default: 20): Number of iterations.
    \item \texttt{decay\_factor} (float, default: 1.0): Momentum accumulation factor $\mu$.
\end{itemize}

\subsection{TI-DIM}

\textbf{Class:} \texttt{TIDIMDeIdentifier} \\
\textbf{Method name:} \texttt{tidim}~\cite{dong2019evading} \\

\noindent\textbf{Parameters:}
\begin{itemize}[noitemsep]
    \item \texttt{epsilon} (float, default: $8/255$): Perturbation budget.
    \item \texttt{alpha} (float, default: $2/255$): Step size.
    \item \texttt{num\_iter} (int, default: 20): Number of iterations.
    \item \texttt{decay\_factor} (float, default: 1.0): Momentum factor.
    \item \texttt{kernel\_size} (int, default: 15): Gaussian kernel size for translation invariance.
    \item \texttt{prob} (float, default: 0.7): Probability of applying diverse input transformation.
\end{itemize}

\subsection{TIP-IM}

\textbf{Class:} \texttt{TIPIMDeIdentifier} \\
\textbf{Method name:} \texttt{tipim}~\cite{yang2021towards} \\

\noindent\textbf{Parameters:}
\begin{itemize}[noitemsep]
    \item \texttt{epsilon} (float, default: $16/255$): Perturbation budget.
    \item \texttt{alpha} (float, default: $2/255$): Step size.
    \item \texttt{num\_iter} (int, default: 100): Number of iterations.
    \item \texttt{gamma} (float, default: 10.0): Perceptual loss weight.
    \item \texttt{decay\_factor} (float, default: 1.0): Momentum factor.
    \item \texttt{kernel\_size} (int, default: 15): Gaussian kernel size.
    \item \texttt{use\_diverse\_input} (bool, default: True): Enable diverse input transformation.
\end{itemize}

\subsection{Chameleon}

\textbf{Class:} \texttt{ChameleonDeIdentifier} \\
\textbf{Method name:} \texttt{chameleon}~\cite{chow2024personalized} \\

\noindent\textbf{Parameters:}
\begin{itemize}[noitemsep]
    \item \texttt{epsilon} (float, default: $16/255$): Perturbation budget.
    \item \texttt{alpha} (float, default: 0.001): Learning rate.
    \item \texttt{num\_iter} (int, default: 20): Number of iterations.
    \item \texttt{lambda\_dsim} (float, default: 1.0): Initial dissimilarity constraint weight.
\end{itemize}

\section{Generative De-Identification Methods}
\label{sec:generative}

Generative methods employ generative models, including GANs, autoencoders, and diffusion models, to synthesize de-identified faces.
These methods generally achieve higher visual quality but require pretrained model weights.
All generative methods are implemented in \texttt{core/fdeid/generative/}.

\subsection{CIAGAN}

\textbf{Class:} \texttt{CIAGANDeIdentifier} \\
\textbf{Method name:} \texttt{ciagan}~\cite{maximov2020ciagan} \\

\noindent\textbf{Parameters:}
\begin{itemize}[noitemsep]
    \item \texttt{weights\_path}: Path to CIAGAN generator weights.
    \item \texttt{retinaface\_path}: Path to RetinaFace detector weights.
    \item \texttt{dlib\_path}: Path to Dlib landmark predictor.
\end{itemize}

\subsection{AMT-GAN}

\textbf{Class:} \texttt{AMTGANDeIdentifier} \\
\textbf{Method name:} \texttt{amtgan}~\cite{hu2022protecting} \\

\noindent\textbf{Parameters:}
\begin{itemize}[noitemsep]
    \item \texttt{weights\_path}: Path to generator weights.
    \item \texttt{reference\_path}: Path to a single reference makeup image.
    \item \texttt{reference\_dir}: Directory with multiple reference images.
    \item \texttt{dlib\_path}: Optional Dlib predictor for facial masking.
\end{itemize}

\subsection{Adv-Makeup}

\textbf{Class:} \texttt{AdvMakeupDeIdentifier} \\
\textbf{Method name:} \texttt{advmakeup}~\cite{yin2021adv} \\

\noindent\textbf{Parameters:}
\begin{itemize}[noitemsep]
    \item \texttt{weights\_path}: Path to encoder/decoder weights directory.
    \item \texttt{target\_name} (str, default: \texttt{00288}): Target identity style.
    \item \texttt{dlib\_path}: Path to Dlib landmark predictor.
\end{itemize}

\subsection{WeakenDiff}

\textbf{Class:} \texttt{WeakenDiffDeIdentifier} \\
\textbf{Method name:} \texttt{weakendiff}~\cite{salar2025enhancing} \\

\noindent\textbf{Parameters and presets:}
\begin{itemize}[noitemsep]
    \item \texttt{prot\_steps}: Protection optimization steps.
    \item \texttt{diffusion\_steps}: Number of diffusion iterations.
    \item \texttt{start\_step}: Diffusion start step.
    \item \texttt{null\_optimization\_steps}: Null text optimization iterations.
    \item \texttt{adv\_optim\_weight}: Adversarial optimization weight.
    \item \texttt{preset} (str): Quality preset:
    \begin{itemize}[noitemsep]
        \item \texttt{fast}: steps=(10, 10, 5)
        \item \texttt{balanced}: steps=(15, 15, 10)
        \item \texttt{quality}: steps=(30, 20, 20)
    \end{itemize}
\end{itemize}

\subsection{DeID-rPPG}

\textbf{Class:} \texttt{DeIDrPPGDeIdentifier} \\
\textbf{Method name:} \texttt{deid\_rppg}~\cite{savic2023identification}  \\

\noindent\textbf{Parameters:}
\begin{itemize}[noitemsep]
    \item \texttt{chunk\_size} (int, default: 64): Number of frames per chunk.
    \item \texttt{face\_size} (int, default: 128): Face crop resolution.
    \item \texttt{overlap} (int, default: 0): Frame overlap between chunks.
\end{itemize}

\subsection{G$^{2}$Face}

\textbf{Class:} \texttt{G$^{2}$FaceDeIdentifier} \\
\textbf{Method name:} \texttt{G$^{2}$Face}~\cite{yang2024g}  \\

\noindent\textbf{Parameters:}
\begin{itemize}[noitemsep]
    \item \texttt{image\_size} (int, default: 256): Input resolution.
    \item \texttt{weights\_path}: Path to G$^{2}$Face checkpoint.
    \item \texttt{arcface\_path}: Path to ArcFace weights.
    \item \texttt{recon\_path}: Path to 3D reconstruction model.
\end{itemize}